%% file: main.tex
\documentclass[10pt,twocolumn,letterpaper]{article}

\input{00_packages}

\threedvfinalcopy
\def\threedvPaperID{285} %

\ifthreedvfinal\fi

\input{00_definitions}

\begin{document}

\title{\ourTitle}
\input{00_authors}

\input{paper_FIG_01_Teaser_COL2}
\thispagestyle{empty} 

\input{00_abstract}                     
\input{01_introduction}					
\input{02_related}

\input{03_method_00}

\input{04_experiments_00}

\input{05_conclusion}
\input{06_disclosure_acks}

{
\small
\bibliographystyle{ieee_fullname}
\bibliography{00_BIB}
\balance
}

\input{sup_mat}

\end{document}

%% file: 00_packages.tex
\usepackage{multirow}
 \usepackage{tabularx}
\usepackage{booktabs}
\usepackage[numbers,sort,compress]{natbib}
\usepackage[dvipsnames]{xcolor}

\usepackage{3dv_PAPER}
\usepackage{times}
\usepackage{epsfig}
\usepackage{graphicx}
\usepackage{amsmath}
\usepackage{amssymb}

\usepackage{cuted}
\usepackage{capt-of}
\usepackage{float}

\usepackage[pagebackref=true,breaklinks=true,letterpaper=true,colorlinks,bookmarks=false]{hyperref}

\usepackage{bm}
\usepackage{xspace}
\usepackage{caption}
\usepackage{subcaption}
\usepackage{balance}
\usepackage[toc,page]{appendix}
\usepackage[normalem]{ulem} %

%% file: 00_definitions.tex
\usepackage[numbers,sort,compress]{natbib}

\usepackage{pifont}

\newcommand{\qheading}[1]{\noindent\textbf{#1}}

\newcommand{\highlight}[1]{\xspace{\color{black} #1}\xspace}
\newcommand{\highlightSYMBOLS}[1]{\xspace{\color{black} #1}\xspace}
\newcommand{\highlightCLAIM}[1]{\xspace{\color{black} #1}\xspace}

\newcommand{\cameraReady}[1]{\xspace{\color{black} #1}\xspace}

\renewcommand{\highlightSYMBOLS}[1]{\xspace{\color{black} #1}\xspace}
\renewcommand{\highlightCLAIM}[1]{\xspace{\color{blue} #1}\xspace}

\renewcommand{\highlightCLAIM}[1]{\xspace{\color{black} #1}\xspace}

\newcommand{\resubsupmat}[1]{\xspace{\color{black} #1}}
\newcommand{\tdv}[1]{\xspace{\color{red} #1}}

\renewcommand{\tdv}[1]{\xspace{\color{black} #1}}

\newcommand{\render}{\highlightSYMBOLS{r}}
\newcommand{\bpart}{\highlightSYMBOLS{p}}
\newcommand{\body}{\highlightSYMBOLS{b}}
\newcommand{\face}{\highlightSYMBOLS{f}}
\newcommand{\hand}{\highlightSYMBOLS{h}}

\newcommand{\fdetail}{\highlightSYMBOLS{d}}
\newcommand{\stripped}{}%

\newcommand{\paalignment}{\highlightSYMBOLS{PA}}
\newcommand{\tralignment}{\highlightSYMBOLS{TR}}

\newcommand{\fused}{\highlightSYMBOLS{\text{fused}}}
\newcommand{\regressorFaceFused}{\regressor_{\face}^{\fused}}
\newcommand{\regressorHandFused}{\regressor_{\hand}^{\fused}}

\newcommand{\ourrefcolor}{\textcolor{black}}

\newcommand{\tab}[1]{\mbox{\ourrefcolor{{Tab.}}~\ref{#1}}}
\newcommand{\fig}[1]{\mbox{\ourrefcolor{{Fig.}}~\ref{#1}}}
\newcommand{\sect}[1]{\mbox{\ourrefcolor{{Sec.}}~\ref{#1}}}

\newcommand{\Tab}[1]{\mbox{\ourrefcolor{{Table}}~\ref{#1}}}
\newcommand{\Fig}[1]{\mbox{\ourrefcolor{{Figure}}~\ref{#1}}}

\newcommand{\modelname}{PIXIE\xspace}
\newcommand{\longmodelname}{Pixels to Individuals: eXpressive Image-based Estimation\xspace}

\newcommand{\moderatorWord}{{moderator}\xspace}

\newcommand{\expose}{\mbox{ExPose}\xspace}
\newcommand{\deca}{\mbox{\mbox{DECA}}\xspace}
\newcommand{\frankmocap}{\mbox{FrankMocap}\xspace}
\newcommand{\websiteURL}{\mbox{\href{https://pixie.is.tue.mpg.de}{\ttfamily pixie.is.tue.mpg.de}}\xspace}

\newcommand{\pare}{\mbox{PARE}\xspace}

\newcommand{\ourTitle}{Collaborative Regression of Expressive Bodies using Moderation}

\newcommand{\na}{\mbox{N/A}\xspace}

\newcommand{\hmr}{\mbox{HMR}\xspace}

\newcommand{\spin}{\mbox{SPIN}\xspace}

\newcommand{\smplifyx}{\mbox{SMPLify-X}\xspace}
\newcommand{\smplifyX}{\smplifyx}
\newcommand{\groundtruth}{\mbox{ground-truth}\xspace}

\newcommand{\stateoftheart}{\mbox{state-of-the-art}\xspace}

\newcommand{\freihand}{\mbox{FreiHAND}\xspace}

\newcommand{\mofa}{\mbox{MoFA}\xspace}
\newcommand{\agora}{\mbox{AGORA}\xspace}

\newcommand{\adam}{\mbox{Adam}\xspace}
\newcommand{\ghum}{\mbox{GHUM}\xspace}

\newcommand{\smpl}{\mbox{SMPL}\xspace}
\newcommand{\smplH}{\mbox{SMPL+H}\xspace}
\newcommand{\smplx}{\mbox{SMPL-X}\xspace}
\newcommand{\smplX}{\smplx}

\newcommand{\smplorsmplx}{\mbox{\highlightSYMBOLS{SMPL(-X)}}\xspace}

\newcommand{\mano}{\mbox{MANO}\xspace}
\newcommand{\flame}{\mbox{FLAME}\xspace}

\newcommand{\caesar}{\mbox{CAESAR}\xspace}

\newcommand{\ringnet}{\mbox{RingNet}\xspace}

\newcommand{\pytorch}{\mbox{PyTorch}\xspace}

\newcommand{\inthewild}{\mbox{in-the-wild}\xspace}
\newcommand{\twoD}{2D\xspace}
\newcommand{\threeD}{3D\xspace}

\newcommand{\supmat}{{Sup.~Mat.}\xspace}

\newcommand{\now}{\mbox{NoW}\xspace}
\newcommand{\ehf}{\mbox{EHF}\xspace}

\newcommand{\NOW}{\mbox{NoW}\xspace}
\newcommand{\threedpw}{\mbox{3DPW}\xspace}
\newcommand{\threeDPW}{\threedpw}

\newcommand{\mtc}{\mbox{MTC}\xspace}

\newcommand{\mpjpe}{\mbox{\highlightSYMBOLS{MPJPE}}\xspace}
\newcommand{\VtoV}{\mbox{\highlightSYMBOLS{V2V}}\xspace}
\newcommand{\PtoS}{\mbox{\highlightSYMBOLS{P2S}}\xspace}

\newcommand{\mask}{S}
\newcommand{\mesh}{M}
\newcommand{\shape}{\bm{\beta}}
\newcommand{\pose}{\bm{\theta}}
\newcommand{\expression}{\bm{\psi}}
\newcommand{\cam}{C}

\renewcommand{\etc}{\mbox{etc}\xspace}
\renewcommand{\etal}{\mbox{et al.}\xspace}
\renewcommand{\ie}{\mbox{i.e.}\xspace}
\renewcommand{\eg}{\mbox{e.g.}\xspace}
\renewcommand{\wrt}{\mbox{w.r.t.}\xspace}

\newcommand{\norm}[1]{\left\lVert#1\right\rVert}
\newcommand{\absval}[1]{\left\lvert#1\right\rvert}
\newcommand{\normabs}[1]{\left\lVert#1\right\rVert_1}
\newcommand{\normmse}[1]{\left\lVert#1\right\rVert_2^2}
\newcommand{\ltwo}[1]{\left\lVert#1\right\rVert_2}
\newcommand{\scalarltwo}[1]{\left\lvert#1\right\rvert^2}

\newcommand{\rgb}{\mbox{RGB}\xspace}
\newcommand{\rgbD}{\mbox{RGB-D}\xspace}

\newcommand{\lighting}{\textbf{l}}

\newcommand{\albedocoeffs}{\boldsymbol{\alpha}}

\newcommand{\encoder}{E}
\newcommand{\moderator}{\mathcal{M}}
\newcommand{\extractor}{\mathcal{L}}
\newcommand{\regressor}{\mathcal{R}}

\newcommand{\bmi}{\mbox{BMI}\xspace}

\newcommand{\includegraphicstrimW}[3]{%
  \begingroup
    \edef\x{\endgroup\noexpand\includegraphics[trim={#2}, clip=true, width=#3]{#1}%
    }%
   \x%
}

\newcommand{\graphsize}{}

%% file: 00_authors.tex
\author{
Yao Feng\textsuperscript{*}
\quad Vasileios Choutas\textsuperscript{*}
\quad
Timo Bolkart \quad Dimitrios Tzionas \quad Michael J. Black\\
Max Planck Institute for Intelligent Systems, T{\"u}bingen, Germany\\
{\tt\small \{yfeng, vchoutas, tbolkart, dtzionas, black\}@tuebingen.mpg.de}\\
{\footnotesize * Equal contribution}
\vspace{-0.7em}
}

%% file: paper_FIG_01_Teaser_COL2.tex
\twocolumn[{
    \renewcommand\twocolumn[1][]{#1}
    \maketitle
    \centering
    \vspace{-2.0em}
    \begin{minipage}{1.00\textwidth}
    \centering%
      \includegraphics[trim=000mm 000mm 000mm 004mm, clip=true, width=1.00 \linewidth]{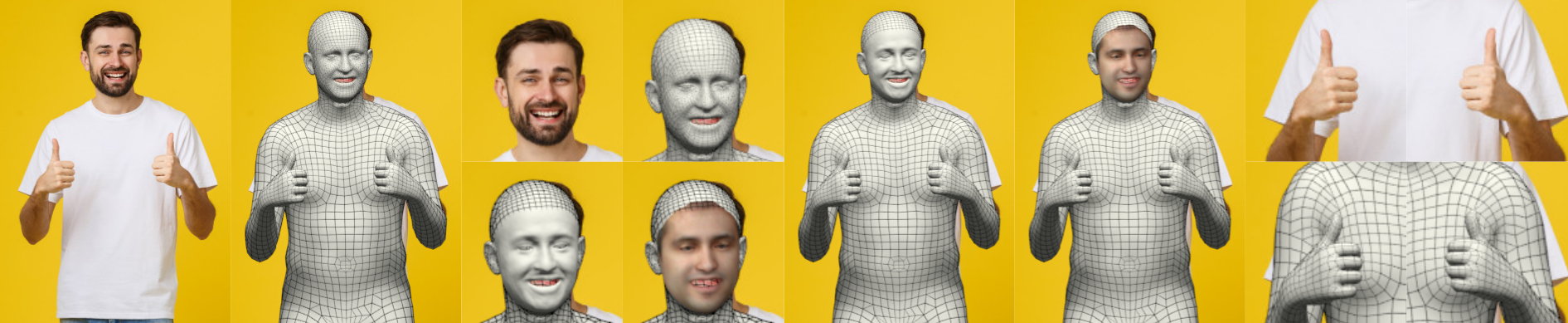}
    \end{minipage}
	\centerline{
	\makebox[0.13\textwidth]{a}
	\makebox[0.13\textwidth]{b}
	\makebox[0.11\textwidth]{c}
	\makebox[0.11\textwidth]{d}
	\makebox[0.13\textwidth]{e}
	\makebox[0.13\textwidth]{f}
	\makebox[0.11\textwidth]{g}
	\makebox[0.11\textwidth]{h}}
    \vspace{-2.0 em}
    \captionof{figure}{
        \textbf{\modelname} estimates expressive \threeD humans (b, e, f) from an \rgb image (a).
        For this, it employs experts for the body, face (c, d), and hands (g, h), which are combined (b, e, f) by a novel \moderatorWord, according to their confidence (see \fig{fig:experts_confidence}). 
        \modelname estimates appropriate body shapes (b) by implicitly learning to reason about gender from an image.
        Finally, \modelname estimates fine facial details, \ie~\threeD surface displacements (c) and albedo (d), similar to \stateoftheart face-only methods.
    }
    \label{fig:teaser}
    \vspace*{+1.5 em}
}]

%% file: 00_abstract.tex
\begin{abstract}
\vspace{-1.0 em}
Recovering expressive humans from images is essential for understanding human behavior.
Methods that estimate \threeD bodies, faces, or hands have progressed significantly, yet separately.
Face methods recover accurate \threeD shape and geometric details, but need a tight crop  and struggle with extreme views and low resolution.
Whole-body methods are robust to a wide range of poses and resolutions, but provide only a rough \threeD face shape without details like wrinkles.
To get the best of both worlds, we introduce \modelname, which produces animatable, whole-body \threeD avatars with realistic facial detail, from a single image.
For this,
\modelname uses two key observations. 
First,
existing work combines independent estimates from body, face,
and hand experts, by trusting them equally.
\modelname introduces a novel \moderatorWord that merges the features of the experts, weighted by their confidence.
\cameraReady{All} part experts can contribute to the whole, using \smplx's shared shape space across all body parts.
Second, human shape is highly correlated with gender, but existing work ignores this.
We label training images as male, female, or non-binary, and train \modelname to infer ``gendered'' \threeD body shapes with a novel shape loss.
In addition to \threeD body pose and shape parameters, \modelname estimates expression, illumination, albedo and \threeD 
facial surface displacements. %
Quantitative and qualitative evaluation shows that \modelname estimates %
\cameraReady{more accurate whole-body shape
and detailed face shape than the state of the art.}
\cameraReady{Models and code are available at \websiteURL}.
\end{abstract}
\vspace{-1.0em}

%% file: 01_introduction.tex
\section{Introduction}

To model human behavior, we need to capture how people look, how they feel, and how they interact with each other.
To facilitate this, our goal is to reconstruct whole-body \threeD shape and pose, facial expressions, and hand gestures from an \rgb image.
This is challenging, as humans vary in shape and appearance, they are highly articulated, they wear complex clothing, they are often occluded, and their face and hands are small, yet highly deformable.
For these reasons, the community studies the
body 	\cite{bogo2016keep,Kanazawa2018_hmr,Kolotouros2019_spin},
hands 	\cite{Boukhayma2019,Ge2019,hasson_2019_cvpr,Zhang_2019_ICCV} and
face 	\cite{Egger_3DMM_survey}
mostly separately.

Recent whole-body statistical models \cite{Joo2018_adam,Pavlakos2019_smplifyx,Xu_2020_CVPR} 
enable approaches to address
the problem holistically, by jointly capturing the body, face and hands. %
\expose \cite{Choutas2020_expose} reconstructs %
\smplX \cite{Pavlakos2019_smplifyx} meshes from an \rgb image, using ``expert'' sub-networks for the body, face and hands.
\cameraReady{%
However, \expose's part experts operate completely independently, as they only
``see" their respective part image. Thus, they do not
exploit the correlations between parts to overcome
challenges like occlusion or motion blur.
}

Face-only methods \cite{DECA_2020,FaceScape_2020} are well studied and recover accurate facial shape, albedo and geometric details, which are important to capture emotions.
However, they need a tight crop around the face and struggle with extreme viewing angles and faces that are small, low-resolution or occluded.
While whole-body methods \cite{Choutas2020_expose,Joo2018_adam,Pavlakos2019_smplifyx,rong2020frankmocap,Xu_2020_CVPR} handle these challenges well, they estimate average-looking face shapes,
\cameraReady{without face albedo and fine geometric details}.

To get the best of all worlds, we introduce {\modelname} (``\longmodelname'').
\modelname estimates expressive whole-bodied \threeD humans from an \rgb image more realistically than existing work. %
To do so, it pushes the state of the art in three ways.

First, \modelname learns not only experts for the body, face and hands, but also a novel moderator that estimates their confidence in each sub-image, and fuses their features weighted by this.
The learned fusion helps improve whole-body shape, using \smplX's shared shape space across all body parts.
Moreover, it helps to robustly estimate head and hand pose when these are ambiguous (\eg occlusions or blur) by 
using
full-body context; 
see \fig{fig:experts_confidence} for examples.

Second, \modelname significantly improves ``gendered'' body shape realism.
While human shape is highly correlated with gender, existing work ignores this and
estimates inaccurate body shapes -- often with the wrong gender or with a gender-neutral shape.
An exception is \smplifyX, but it uses an offline gender classifier and fits a gender-specific \smplX model.
Instead, using a single unisex \smplX model enables end-to-end training of neural nets.
\modelname adopts this approach, and learns to implicitly reason about shape.
For 
this, we define male, female, and non-binary body-shape priors within the \smplX shape space.
At training time, given automatically created gender labels for input images, we train \modelname to output plausible shape parameters for the specified gender.
At inference time, \modelname needs no gender labels, is applicable to any \inthewild image, and supports non-binary genders.
Note that this approach is general and is relevant for the broader community (face, body, whole-body). %
\tdv{Body shape is also
correlated with face shape \cite{gallucci2020prediction,gunel2019face,kocabey2017face2BMI}.}
Thus, we do the same ``gendered'' training for our face expert; this allows \modelname to %
\highlight{use face information to inform body shape}.
This training and network architecture \highlightCLAIM{significantly} improves body shape both qualitatively and quantitatively.

\input{paper_FIG_02_Confidence}

Third, \modelname's face expert additionally infers facial albedo and dense \threeD facial-surface displacements.
For this, we draw inspiration from Feng \etal~\cite{DECA_2020}, and go beyond them in three ways:
(1)     We use a whole-body shape space, rather than a face-only space, to capture correlations between the body and face shape.
(2)     We use photometric and identity losses on faces \highlight{to inform whole-body shape}.
(3)     We use the inferred geometric details only when the face expert is confident, as judged by the moderator.
As shown in \fig{fig:teaser}, this results in whole-body \threeD humans with detailed faces that can be fully animated.

To summarize, here we make three key contributions:
\mbox{(1) 	We train} a novel \moderatorWord, that infers the confidence of body-part experts and fuses their features weighted by this. This improves shape and pose inference under ambiguities.
\mbox{(2) 	We train} the network to implicitly reason about gender, \ie without gender labels at test time, with a novel ``gendered'' \threeD shape loss that encourages likely body shapes.
\mbox{(3) 	We extend} our face expert with branches that estimate facial albedo and \threeD facial-surface displacements, enabling whole-body animation with a realistic face.
\modelname is a step towards automatic, accurate and realistic \threeD avatar creation from a single \rgb image.
\cameraReady{Models and code are available for research purposes mat \websiteURL.}

%% file: paper_FIG_02_Confidence.tex
\begin{figure}
    \centering
    \includegraphics[trim=000mm 000mm 000mm 000mm, clip=true, width=0.99\linewidth]{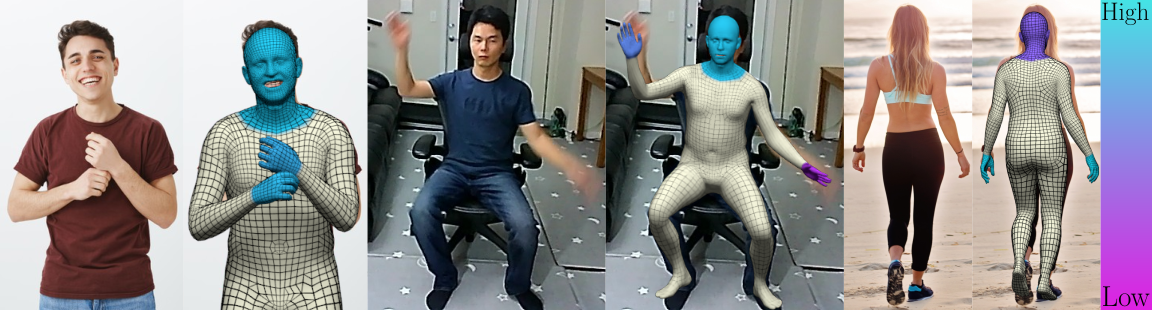}
     \vspace{-0.7 em}
    \caption{
        \modelname infers the confidence of its body, face and hand experts, and fuses their features accordingly.
        Challenges, like occlusions, are resolved with full-body context.
        (L) Input image.
        (R) Color-coded part-expert confidence.
    }
    \vspace{-1.6em}
    \label{fig:experts_confidence}
\end{figure}

%% file: 02_related.tex
\section{Related work}      \label{sec:related}

\textbf{Body reconstruction:}
For years, the community focused on the prediction of \twoD or \threeD
landmarks for the body \cite{Cao2018_openpose}, face \cite{bulat2017far}
and hands \cite{simon2017hand,Wang:2018:MCC}, with a recent shift towards estimating 
\threeD model parameters \tdv{\cite{bogo2016keep,joo2020eft,Kanazawa2018_hmr,omran2018neural,pare,pavlakos2018learning,romp}} or
\threeD surfaces \tdv{\cite{kolotouros2019convolutional,lin2021end-to-end,Saito_2019_ICCV,Saito_2020_CVPR,varol2018bodynet}}.
One line of work simplifies the problem by using proxy representations like
\twoD joints 			\cite{bogo2016keep,grauman2003inferring,guan_iccv_scape_2009,MuVS_3DV_2017,martinez_2017_3dbaseline,pavlakos2018learning,sigal2006predicting,tome2017lifting,zhao_2019_cvpr_semgcn},
silhouettes 			\cite{agarwal_trigs_3d_poses,MuVS_3DV_2017,pavlakos2018learning},
part labels 			\cite{omran2018neural,Ruegg:AAAI:2020} or
dense correspondences 	\cite{yu_iccv2019_hybrid,zeng20203d}.
\cameraReady{These are then ``lifted'' to \threeD, either as part of an energy term}
\cite{bogo2016keep,MuVS_3DV_2017,zanfir_2018_cvpr} or using a regressor \cite{martinez_2017_3dbaseline,omran2018neural,pavlakos2018learning,tome2017lifting}.
To overcome ambiguities, they use priors such as known limb lengths
\cite{lee1985determination}, %
\tdv{joint angle limits}
\cite{akhter2015pose}, or a statistical body model \cite{bogo2016keep,MuVS_3DV_2017,omran2018neural,Pavlakos2019_smplifyx,pavlakos2018learning} like \smpl \cite{SMPL:2015} or \smplx \cite{Pavlakos2019_smplifyx}.
While these approaches benefit from \twoD annotations,
they cannot overcome errors in the proxy features and do not fully exploit image context.
The alternative is to directly regress \tdv{%
\threeD skeletons 				\cite{li2015maximum,pavlakos2017coarse,sun2017compositional,sun2018integral,bugra_bmvc_2016},
statistical model parameters 	
\tdv{\cite{Choutas2020_expose,Fieraru2020,joo2020eft,Kanazawa2018_hmr,kanazawa_2019_cvpr,Kolotouros2019_spin,pare,romp}}, 
\threeD meshes 					\tdv{\cite{kolotouros2019convolutional,lin2021end-to-end}},
depth maps \cite{Gabeur_2019_ICCV,Smith_2019_ICCV},
\threeD voxels \cite{varol2018bodynet,Zheng_2019_ICCV} or
\tdv{distance fields \cite{Saito_2019_ICCV,Saito_2020_CVPR}}
 from the image pixels.
}

\textbf{Face reconstruction:}
Most modern monocular \threeD face reconstruction methods estimate the parameters of a pre-computed statistical face model~\cite{Egger_3DMM_survey}.
Similar to the body literature, this problem is tackled with both optimization \cite{AldrianSmith2013,Bas2017fitting,VetterBlanz1998,Thies2016} and regression methods \cite{Feng2018,Jackson2017,Sanyal2019_ringnet,Tewari2018}.
Many learning-based approaches follow an analysis-by-synthesis strategy~\cite{Deng2019,Tewari2018,Tewari2017}, which jointly estimates geometry, albedo, and lighting, to render a synthetic image \cite{Loper:ECCV:2014,ravi2020pytorch3d} that is compared with the input.
Recent work \cite{DECA_2020,Deng2019,Genova2018} further employs face-recognition terms \cite{Cao2018_VGGFace2} during training to reconstruct more accurate facial geometry.
Even geometric details, such as wrinkles, can be learned from large collections of in-the-wild images \cite{DECA_2020, LuanTran2019}.
We refer to Egger \etal~\cite{Egger_3DMM_survey} for a comprehensive overview.
The major downsides of face-specific approaches are their need for tightly cropped face images and their inability to handle non-frontal images.
The latter is mainly due to the lack of supervision; \twoD landmarks may be missing \cameraReady{or the face might not even be detected at all,}
in which case the photometric term is not applicable. By
integrating face and body regression,
\modelname regresses head pose and shape robustly in situations where face-only methods fail 
\highlightCLAIM{and lets the face contribute to whole-body shape estimation.}

\textbf{Hand reconstruction:}
While hand pose estimation is most often performed from \rgbD data, there has been a recent shift towards the use of monocular \rgb images \cite{Baek_2019_CVPR,Boukhayma2019,honnotate2020,hasson_2019_cvpr,Iqbal_2018_ECCV,kulon2019rec,Mueller_2018_GANerated,Tekin_2019_CVPR,brox_ICCV_2017}.
Similar to the body, we split these into methods that predict
\threeD joints 							\cite{Iqbal_2018_ECCV,Mueller_2018_GANerated,Tekin_2019_CVPR,brox_ICCV_2017},
parameters of a statistical hand model 	\cite{Baek_2019_CVPR,Boukhayma2019,hasson_2019_cvpr,kulon2019rec,Zhang_2019_ICCV}, such as \mano \cite{romero2017embodied}, or a
\threeD surface 							\cite{Ge2019,Kulon_2020_CVPR}.

\textbf{Whole-body reconstruction:}
Recent methods approach the problem of human reconstruction holistically. 
Some of these estimate \threeD landmarks for the body, face and hands \cite{jin2020whole,Weinzaepfel2020_dope}, but not their \threeD surface.
This is addressed by whole-body statistical models \cite{Joo2018_adam,Pavlakos2019_smplifyx,Xu_2020_CVPR},
that jointly capture the \threeD surface for the body, face and hands.

\smplifyx~\cite{Pavlakos2019_smplifyx} fits \smplx~\cite{Pavlakos2019_smplifyx} to \twoD body, hand and face keypoints \cite{Cao2018_openpose} estimated in an image. 
Xiang \etal~\cite{Xiang2019} estimate both \twoD keypoints and a part orientation field and fit \adam~\cite{Joo2018_adam} to these.
Xu \etal~\cite{Xu_2020_CVPR} fit \ghum~\cite{Xu_2020_CVPR} to detected body-part image regions.
While these methods work, they are based on optimization, consequently they are slow and do not scale up to large datasets.

Deep-learning methods \cite{Choutas2020_expose,rong2020frankmocap} tackle these limitations, and quickly regress \smplx parameters from an image.
\expose~\cite{Choutas2020_expose} uses ``expert'' sub-networks for the body, face and hands; 
\tdv{the body expert estimates the body and rough part (hand/face) pose from the full-body image, 
while part experts refine the rough part poses using only 
local image information (hand/face crop).}
\expose merges the output of its experts by always trusting them. 
Instead, we evaluate the confidence of each expert for each sub-image
and \cameraReady{%
fuse body/face and body/hand features weighted by this.}
\cameraReady{%
To account for different body-part sizes,
we use
\expose's body-driven attention, and multiple data sources for both part-only and whole-body supervision.
}

\frankmocap~\cite{rong2020frankmocap} 
is similar to \expose
and adds an (optional) 
optimization step to better align 
the estimated \smplX mesh with the image. 
Zhou \etal~\cite{zhou2021pixieLike} train a network to regress 
a body-and-hands (\smplH) model \cite{romero2017embodied} and the \tdv{detailed} \mofa \cite{Tewari2017} face model
from an \rgb image,
following a body-part attention mechanism and multi-source training like \expose.
Note that %
\smplH and \mofa
are disparate models, which are (offline) manually cut-and-stitched together.
Instead, we use the whole-body \smplX model \cite{Pavlakos2019_smplifyx} that  captures the shape of all body parts together, thus no stitching is required.
Zhou \etal fuse only hand-body features in a ``binary'' fashion, while their face model is ``disconnected'' from the body.
Instead, we fuse both face-body and hand-body features in a ``fully analog'' fusion, 
\highlightCLAIM{and thus our face expert can %
inform the whole-body shape.} 
Zhou \etal have no face camera, and need PnP-RANSAC \cite{PnpRansac} and Procrustes to align their face to the image.
Instead, we infer a face-specific camera and need no extra steps. 
Zhou \etal use a complicated architecture, with several modules that are trained separately, and is applicable only to whole bodies. 
Instead, we use no intermediate tasks to avoid possible sources of error and train our model end to end.
Our full model is applicable to whole bodies, but the part experts are also (separately) applicable to part-only data.

%% file: 03_method_00.tex
\section{Method}		\label{sec:method}

Here we introduce \modelname, a novel model for reconstructing \smplX~\cite{Pavlakos2019_smplifyx} humans with a realistic face from a single \rgb image.
It uses a set of expert sub-networks for body, face/head, and hand regression, and combines them in a bigger network architecture with three main novelties:
(1) We use a novel moderator that assesses the confidence of part experts and fuses their features weighted by this, for robust inference under ambiguities, like strong occlusions.
(2) We use a novel ``gendered'' shape loss, to improve body shape realism by learning to implicitly reason about gender.
(3) In addition to the albedo predicted by our face expert, we employ the surface details branch of Feng \etal~\cite{DECA_2020}. 

\input{03_method_01_body_model}         \input{paper_FIG_03_ARCH}
\input{03_method_02_architecture}
\input{03_method_03_losses}
\input{03_method_04_impl_details}

%% file: 03_method_01_body_model.tex
\subsection{Expressive \threeD Body Model}	\label{sec:technical_body_model}

We use the expressive \smplx \cite{Pavlakos2019_smplifyx} body model, which captures whole-body pose and shape, including facial expressions and finger articulation.
It is a differentiable function $\mesh(\shape, \pose, \expression)$, parameterized by shape $\shape$, pose $\pose$ and expression $\expression$, that produces a \threeD mesh.
The shape parameters $\shape \in \mathbb{R}^{200}$ are coefficients of a linear shape space, learned from registered \caesar\cite{CAESAR}  scans.
This is a joint shape space for the body, face, and hands, naturally capturing their shape correlations.
The expression parameters $\expression \in \mathbb{R}^{50}$ are also coefficients of a low-dimensional linear space.
The overall pose parameters $\pose$ consist of body, jaw and hand pose vectors.
Each joint rotation is encoded as a 6D vector \cite{zhou_2019_cvpr}, except for the jaw, which uses Euler angles, \ie a \threeD vector.
We follow the notation of~\cite{Kanazawa2018_hmr} and denote posed joints with $X(\pose, \shape) \in \mathbb{R}^{J\times 3}$, where $J=55$.
\qheading{Camera:} To reconstruct \smplX from images, we use the weak-perspective camera model with scale $s \in \mathbb{R}$ and translation $\bm{t} \in \mathbb{R}^2$.
We denote the joints $X$ and model vertices $M$ projected on the image with $\bm{x} \in \mathbb{R}^{J \times 2}$ and $\bm{m} \in \mathbb{R}^{V \times 2}$.
\vspace{-1.3em}

%% file: paper_FIG_03_ARCH.tex
\begin{figure}
    \centering
    \includegraphics[width=1.02\linewidth]{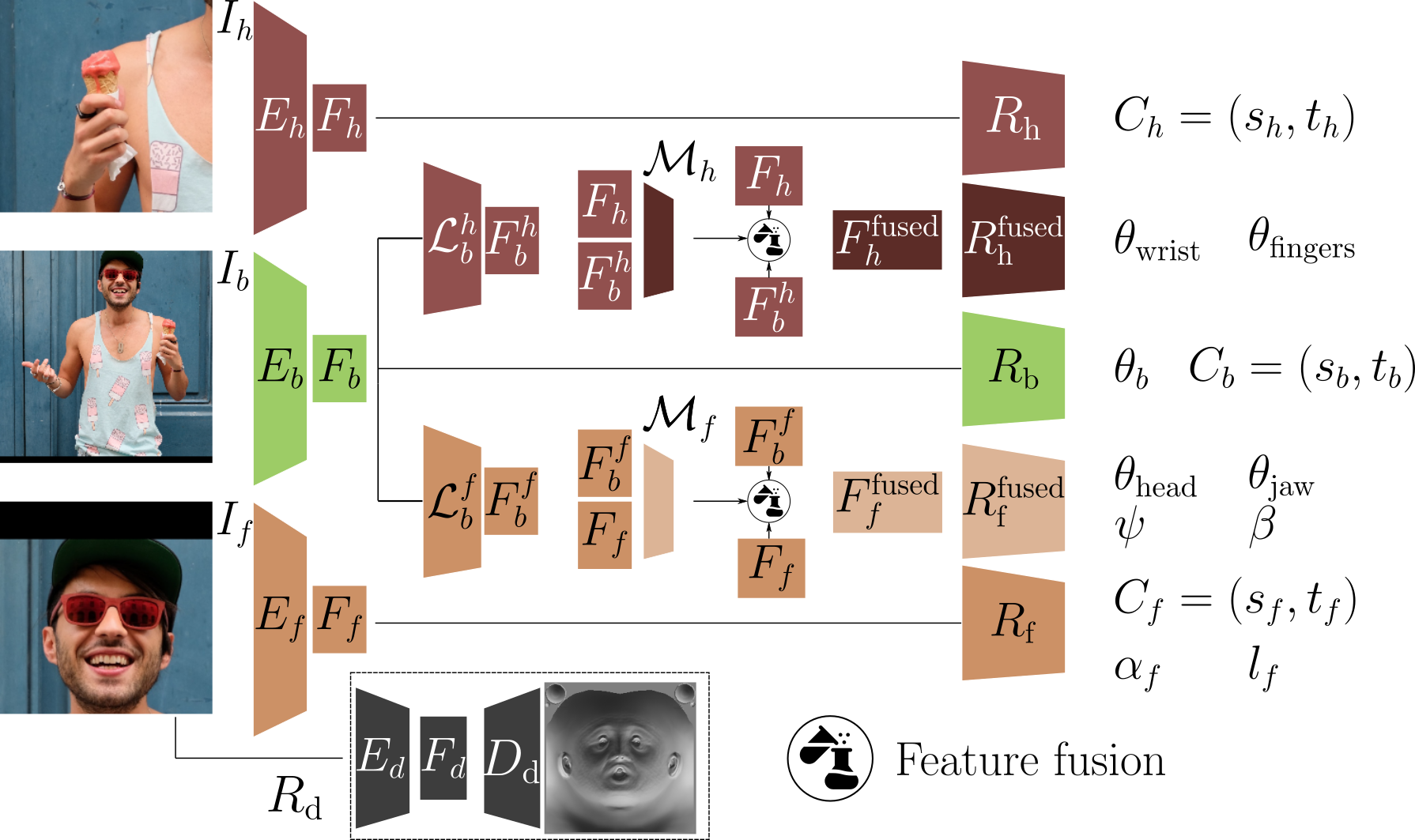}
	\vspace{-1.5 em}
    \caption[]{
		Body, face/head and hand image crops $\{I_{\body}$, $I_{\face}$, $I_{\hand}\}$ are fed to the expert encoders $\{\encoder_{\body}$, $\encoder_{\face}$, $\encoder_{\hand}\}$ to produce part-specific features $\{F_{\body}$, $F_{\face}$, $F_{\hand}\}$. 
		Our novel moderators $\{\moderator_{\face}$, $\moderator_{\hand}\}$ estimate the confidence of experts for these images, and fuse face-body and hand-body features weighted by this, to create $\{ F_{\face}^{\fused}, F_{\hand}^{\fused} \}$. 
        These are fed to $\{ \regressorFaceFused, \regressorHandFused \}$ for robust regression.
        \cameraReady{\deca's \cite{DECA_2020} $R_{\fdetail}$}
        estimates fine geometric details.
        Icon from \href{https://www.flaticon.com/free-icon/mixing_2699568}{Freepik}. 
    }
    \label{fig:arch}
    \vspace{-1.3em}
\end{figure}

%% file: 03_method_02_architecture.tex
\subsection{\modelname Architecture}	        	\label{sec:architecture}

\modelname uses the architecture of \fig{fig:arch}, and is trained end to end.
All model components are described below.

\qheading{Input images:}
Given an image $I$ with full resolution, we assume a bounding box around the body.
We use this to crop and downsample the body to $I_{\body}$ to feed our network.
However, this makes hands and faces too low resolution.
We thus use an attention mechanism \cite{Choutas2020_expose} to extract from $I$ high-resolution crops for the face/head, $I_{\face}$, and hand, $I_{\hand}$.

\qheading{Feature encoding:}
We feed $\{I_{\body}, I_{\face}, I_{\hand}\}$ to separate expert encoders $\{\encoder_{\body}, \encoder_{\face}, \encoder_{\hand}\}$ to extract features $\{F_{\body}, F_{\face}, F_{\hand}\}$.
We use ResNet-50~\cite{he2016deep} for the face/head and hand experts to generate $F_{\face}, F_{\hand} \in \mathcal{R}^{2048}$.
The body expert $\encoder_{\body}$ uses HRNet~\cite{SunXLW19}, followed by convolutional layers that aggregate the multi-scale feature maps, to generate $F_{\body} \in \mathcal{R}^{2048}$.

\qheading{Feature fusion (moderator):}
We identify the expert pairs of \{body, head\} and \{body, hand\} as complementary,
and learn the novel moderators  $\{ \moderator_{\face}, \moderator_{\hand} \}$
that build ``fused'' features   $\{ F_{\face}^{\fused}, F_{\hand}^{\fused} \}$
and feed them to face/head and hand regressors $ \{\regressorFaceFused, \regressorHandFused \}$ (described below) for more informed inference.
A moderator is implemented as a multi-layer perceptron (MLP) and gets the body, $F_{\body}$, and part, $F_{\bpart}$ ($F_{\face}$ or $F_{\hand}$), features and fuses them with a weighted sum:
\begin{align}
    &F_{\bpart}^{\fused}        = w_{\bpart} F_{\body}^{\bpart} + (1 - w_{\bpart}) F_{\bpart},                      \label{eq:mod1}  \\
    &w_{\bpart}  = \frac{1}{1 + \exp\left(  -t * \moderator_{\bpart}(F_{\body}^{\bpart}, F_{\bpart})  \right)},     \label{eq:mod2}  
\end{align}
where
$\moderator_{\bpart}$   ($\moderator_{\face}$ or $\moderator_{\hand}$) is the part moderator,
$w_{\bpart}$            ($w_{\face}$ or $w_{\hand}$) is the expert's confidence, and
$F_{\body}^{\bpart}$    ($F_{\body}^{\face}$ or $F_{\body}^{\hand}$) is the body feature $F_{\body}$ transformed by the respective ``extractor'', \ie the linear layer $\extractor^{\bpart}$ ($\extractor^{\hand}$ or $\extractor^{\face}$) between the body encoder $\encoder_{\body}$ and part moderator $\moderator_{\bpart}$.
Finally, $t$            is a learned temperature weight, jointly trained with all network weights with the losses of \sect{subsec:losses}, with no $t$-specific supervision.

\qheading{Parameter regression:}
We use two main regressor types:
(1)
We use the body, face/head, and hand $\{ \regressor_{\body}, \regressor_{\face}, \regressor_{\hand} \}$ regressors, that get features \emph{only} from the respective expert encoder \{$F_{\body}$, $F_{\face}$, $F_{\hand}$\}.
$\regressor_{\body}$ infers the camera $\cam_{\body} = (s_{\body}, \bm{t}_{\body})$, and body rotation and pose $\pose_{\body}^{\stripped}$
up to (excluding) the head and wrist.
$\regressor_{\face}$ infers the camera $\cam_{\face} = (s_{\face}, \bm{t}_{\face})$, face albedo $\albedocoeffs_{\face}$, and lighting $\lighting_{\face}$.
$\regressor_{\hand}$ infers the camera $\cam_{\hand} = (s_{\hand}, \bm{t}_{\hand})$.
(2)
We use the face/head, $\regressorFaceFused$, and hand, $\regressorHandFused$,
regressors that get from moderators the ``fused'' features, $F_{\face}^{\fused}$ and $F_{\hand}^{\fused}$. 
$\regressorHandFused$ infers the wrist $\pose_{\text{wrist}}$ and finger pose $\pose_{\text{fingers}}$.
$\regressorFaceFused$ infers expressions $\expression$, head rotation $\pose_{\text{head}}$, and jaw pose $\pose_{\text{jaw}}$.
Importantly, $\regressorFaceFused$ also infers body shape $\shape$, \highlight{letting our face expert contribute to whole-body shape.}

\qheading{Detail capture:}
We use the fine geometric details branch $R_{\fdetail}$ of Feng \etal~\cite{DECA_2020} that, given a face image $I_{\face}$, estimates dense \threeD displacements on top of \flame's~\cite{FLAME:SiggraphAsia2017} surface. 
We convert the displacements from \flame's to \smplx's UV map, and apply them on \modelname's inferred head shape. 
However, inferring geometric details from full-body images is not trivial; faces tend to be much noisier in these compared to face-only images.
We account for this with our moderator, and use the inferred displacements only
when the face/head expert is confident. %

%% file: 03_method_03_losses.tex
\subsection{Training Losses}	    \label{subsec:losses}

To train \modelname we use body, hand and face losses:
\begin{align}
     L 	&= L_{\text{body}} + L_{\text{hand}} +
     L_{\text{face}} +
     L_{\text{update}},
\end{align}
defined as follows; the hat (\eg $\hat{\bm{x}}$) denotes ground truth. 

\qheading{Body losses:}
Following \cite{Choutas2020_expose}, we use a combination of a \twoD re-projection, a \threeD joint, and a \smplx parameter loss:
\begin{align}
     L_{\text{body}} &= L_{\text{\twoD/\threeD-Joints}}^{\text{body}} + L_{\text{params}}^{\text{body}}, \label{eq:body_loss}\\
     L_{\text{\twoD/\threeD-Joints}}^{\text{body}} 	&= \sum_{j=1}^J \normabs{\hat{\bm{x}}_j - \bm{x}_j}
                                                     + \sum_{j=1}^J \normabs{\hat{\bm{X}}_j - \bm{X}_j}, 			   \\
     L_{\text{params}}              ^{\text{body}} 	&= \normmse{\hat{\pose} - \pose} + \normmse{\hat{\shape} - \shape}.
\end{align}

\qheading{Hand losses:}
We employ a similar set of losses to train the \threeD hand pose and shape estimation network:
\begin{align}
	L_{\text{hand}} &= L_{\text{\twoD/\threeD-Joints}}^{\text{hand}} + L_{\text{params}}^{\text{hand}},
\end{align}
defined similarly to $L_{\text{\twoD/\threeD-Joints}}^{\text{body}}$ and $L_{\text{params}}^{\text{body}}$ of the body, but using the hand joints and pose parameters $\pose_{\text{wrist}}$ and $\pose_{\text{fingers}}$. 

\qheading{Face losses:}
\tdv{We} adopt standard losses used by the \threeD face estimation community \tdv{\cite{DECA_2020, Deng2019}}:
\begin{align}
L_{\text{face}} &=
L_{\text{lmk}} + L_{\text{lmk-closure}} +
L_{\text{params}}^{\text{face}} +
L_{\text{pho}} +  L_{\text{id}}.
\end{align}
The landmark loss penalizes the difference between \cameraReady{detected \cite{bulat2017far}
target \twoD landmarks} $\hat{\bm{m}_j}$ and respective model landmarks (lying on $\mesh_{\face}$) projected on the image plane, $\bm{m}_j$:
\begin{equation}
	L_{\text{lmk}} = \sum_{j=1}^{N_{\text{lmk}}} \normabs{\hat{\bm{m}}_j - \bm{m}_j}.
\end{equation}
\tdv{Following \cite{DECA_2020},} we also compute a loss for the set $E$ of landmarks on the upper, lower eyelid and upper, lower lip:
\begin{equation}
	L_{\text{lmk-closure}} = \sum\limits_{(i,j) \in E} \normabs{(\hat{\bm{m}}_i - \hat{\bm{m}}_j) - (\bm{m}_i - \bm{m}_j)}.
\end{equation}

The face parameter loss $L_{\text{params}}^{\text{face}}$ follows $L_{\text{params}}^{\text{body}}$, but for face pose $\pose_{\text{\text{face}}}$ only.
This loss is only used for face crops from body data, when the target face pose is available.

Given the predicted \threeD face mesh $\mesh_{\face}$ as a subset of $\mesh$, face albedo $\albedocoeffs_{\face}$ and lighting $\lighting_{\face}$, 
we render a synthetic image $I_{\render}$ for the input subject using the differentiable renderer from Pytorch3D~\cite{ravi2020pytorch3d}.
We then minimize the difference between the input face image $I_{\face}$ and the rendered image $I_{\render}$:
\begin{equation}
	L_{\text{pho}} =
	\norm{\text{\mask}
	\odot (I_{\face} - I_{\render})}_{1,1},
    \label{eq:photometric}
\end{equation}
where $\text{\mask}$ is a binary face mask with value $1$ in the face skin region, and $0$ elsewhere, and $\odot$ denotes the Hadamard product.
The segmentation mask prevents errors from non-face regions  influencing the optimization, and we use the segmentation network of Nirkin \etal~\cite{Nirkin2018} to extract $\text{\mask}$.
The image formation process is the same as in Feng \etal~\cite{DECA_2020}.

Following \cite{Deng2019,gecer2019ganfit}, we use a pre-trained face recognition network~\cite{Cao2018_VGGFace2}, $f_{\text{id}}$, to compute embeddings for the rendered image $I_{\render}$ and the input $I_{\face}$.
We then maximize the cosine similarity between the two identity embeddings
\begin{equation}
	L_{\text{id}} = 1 - \frac{<f_{\text{id}}(I_{\face}),
	f_{\text{id}}(I_{\render})>}{\ltwo{f_{\text{id}}(I_{\face})} \cdot \ltwo{f_{\text{id}}(I_{\render})}}.
\end{equation}

\qheading{Priors:}
Due to the difficulty of the problem, we use additional priors to constrain \modelname to generate plausible solutions.
For expression parameters, we use a Gaussian prior:
\begin{equation}
    L_{\text{exp}}\left(\expression\right) = \normmse{\expression}.
    \label{eq:expression_prior}
\end{equation}
We also add soft regularization on jaw and face pose: 
\begin{align}
    &L_{\text{jaw}}( \pose_{\text{jaw}}) = 
    \scalarltwo{ \pose_{\text{jaw}}^\text{pitch}} 		 + 
    \scalarltwo{\pose_{\text{jaw}}^\text{roll}}           + 
	\scalarltwo{\min(\pose_{\text{jaw}}^\text{yaw}, 0)},  \\
	\label{eq:jaw_pose_prior}
    &L_{\text{face}}(
    \pose_{\text{face}}) = \scalarltwo{\max(%
    \absval{%
    \pose_{\text{face}}^\text{yaw}
    }, 90)
    }. 
\end{align}
All these priors are ``standard'' regularizers, empirically found to discourage implausible configurations (extreme values, unrealistic shape/pose, inter-penetrations, \etc).

\qheading{Gender:}
As gender strongly affects body shape, we use a gender-specific shape prior during training, when gender labels are available.
For this, we register \smplx to \caesar \cite{CAESAR} scans, and compute the mean $\bm{\mu}$ and covariance $\Sigma$ of shape parameters for each gender.
We then use:
\begin{equation}
    L_{\text{shape}}\left(\shape\right) =
	\begin{cases}
		(\shape - \mu_{\text{F}})^T \Sigma_{\text{F}} (\shape - \mu_{\text{F}}) & \text{if female}\\
		(\shape - \mu_{\text{M}})^T \Sigma_{\text{M}} (\shape - \mu_{\text{M}}) & \text{if male} \\
		\normmse{\shape} & \text{o/w}.
	\end{cases}
  \label{eq:shape_prior}
\end{equation}
When gender is unknown, we use a Gaussian prior computed over all scans/registrations, irrespective of gender. 
Please note that we do not need gender labels for inference.

\qheading{Feature update loss:}
We encourage the transformed body features $F_{\body}^{\bpart}$ ($F_{\body}^{\face}$ or $F_{\body}^{\hand}$) to match $F_{\bpart}^{\text{fused}}$ with a loss that was empirically found to stabilize network training:
\begin{equation}
        L_{\text{update}} = \normabs{F_{\body}^{\bpart} - F_{\bpart}^{\fused}}.
\end{equation}

%% file: 03_method_04_impl_details.tex
\subsection{Implementation Details}	\label{subsection:impl}

\qheading{Training data:}
For whole-body data we use the curated \smplx fits of \cite{Choutas2020_expose}, and \smplx fits to whole-body COCO data~\cite{jin2020whole}.
For hand-only data we use \freihand~\cite{Freihand2019} and Total Motion \cite{Xiang2019}.
For face/head data we use \mbox{VGGFace2}~\cite{Cao2018_VGGFace2} and \cameraReady{detect $N_{\text{lmk}}=68$ \twoD landmarks} with the method of Bulat \etal~\cite{bulat2017far}.
We get gender annotations by running the method of Rothe \etal~\cite{Rothe-ICCVW-2015} on many photos per identity and using majority voting to improve robustness.
For data augmentation, see \supmat

\qheading{Network training:}
We do multi-step training that empirically aids stability. 
We pre-train on part-only data, and train on whole-body data end to end; for details see \supmat

%% file: 04_experiments_00.tex
\section{Experiments}

                                        \input{paper_TAB_01_experiments_EHF}
\input{04_experiments_01_datasets}
\input{04_experiments_02_metrics}
\input{04_experiments_03_eval_quant}

\input{04_experiments_04_eval_qual}

\input{04_experiments_05_future}

%% file: paper_TAB_01_experiments_EHF.tex
\newcommand{\optimization}{Optimization}
\newcommand{\regression}{Regression}

\renewcommand{\optimization}{O}
\renewcommand{\regression}{R}

\begin{table*}
\centering
\scriptsize
\resizebox{1.00\linewidth}{!}{
    \centering
    \begin{tabular}{lc|l|c|c|c|c|c|c|c|c|c|c|c|c|c}
        \toprule
        \multirow{2}{*}{Method} & \multirow{2}{*}{Type} &
        \multirow{2}{10ex}{Body model}						    &
        \multirow{2}{*}{Time (s)}						        &
        \multicolumn{4}{c|}{\paalignment-\VtoV   (mm) $\downarrow$} 	    &
        \multicolumn{4}{c|}{\tralignment-\VtoV   (mm) $\downarrow$} 	&
        \multicolumn{2}{c|}{\paalignment-\mpjpe (mm) $\downarrow$} 	    &
        \multicolumn{2}{c}{\paalignment-\PtoS (mm) $\downarrow$}
        \\
        \cline{5-16}
		{} & {}                                     & {}          & {}
		& All           & Body          & L/R hand                    & Face
		& All           & Body          & L/R hand                    & Face
		& MPJPE-14 & L/R hand & Mean & Median\\
		\midrule
		$\text{\smplifyx}^{\prime}$~\cite{Pavlakos2019_smplifyx} & \optimization
		& \smplx
		& ~40-60
		& 52.9 & 56.37         & 11.4/12.6                   & 4.4
		& 79.5 &  92.3   & 21.3/22.1                   & 10.9
		& 73.5          & 12.9/13.2
		& 28.9 & 18.1          \\
		\midrule
        \smplifyx   \cite{Pavlakos2019_smplifyx} & \optimization
        & \smplx
        & ~40-60
        & 65.3          & 75.4          & 11.6/12.9                   & 4.9
        & 93.0          & 116.1          & \textbf{23.8}/\textbf{24.9}                   & \textbf{11.5}
        & 87.6          & 12.2/13.5
        & 36.8          & 23.0          \\
        \mtc        \cite{Xiang2019} & \optimization             & \adam
        & 20
        & \na           & \na           & \na                         & \na
        & \na           & \na           & \na                         & \na
        & 107.8         & 16.3/17.0           & 41.3          & 29.0          \\
		\midrule
        \spin
        \cite{Kolotouros2019_spin} & \regression
        & \smpl
        & \textbf{0.01}
        & \na           & 60.6          & \na                         & \na
        & \na           & 96.8          & \na                         & \na
        & 102.9         & \na
        & 40.8          & 28.7          \\
        \frankmocap \cite{rong2020frankmocap} & \regression
        & \smplx
        & 0.08
        & 57.5          & 52.7 & 12.8/12.4                   & \na
        & 76.9  &         80.1  &         32.1 /        31.9  &   \na
        & 62.3          & 13.2/12.6
        & 31.6          & 19.2          \\
        \expose     \cite{Choutas2020_expose} & \regression
        & \smplx
        & 0.16
        & \textbf{54.5} & \textbf{52.6}          & 13.1/12.5                   & 4.8
        & \textbf{65.7} &         76.8  &         31.2 /        32.4  &         15.9
        & 62.8          & 13.5/12.7
        & \textbf{28.9} & \textbf{18.0} \\
        \textbf{\modelname} (ours) & \regression                        & \smplx
        & 0.08-0.10
        & 55.0          & 53.0          & \textbf{11.2}/\textbf{11.0} & \textbf{4.6}
        & 67.6  & \textbf{75.8} & 25.6/27.0 & 14.2
        & \textbf{61.5} & \textbf{11.7}/\textbf{11.4}
        & 29.9          & 18.4          \\
        \bottomrule
    \end{tabular}
    }
    \vspace{-0.5em}
    \caption{
        Evaluation on \ehf \cite{Pavlakos2019_smplifyx}.
        \modelname is on par with the state of the art \wrt body and face performance,
        but predicts better
        hand poses.
        $\text{\smplifyx}^{\prime}$ uses the \groundtruth focal length (\emph{excluded from bold}).
        Run-times were measured on an Intel Xeon W-2123 3.60GHz machine with a NVIDIA Quadro P5000 GPU.
        ``O/R'' denotes Optimization/Regression.
    }
    \vspace{-1.0em}
    \label{tab:ehf_full}
\end{table*}

%% file: 04_experiments_01_datasets.tex
\subsection{Evaluation Datasets}

\qheading{\ehf~\cite{Pavlakos2019_smplifyx}:}
We evaluate whole-body accuracy on this.
It has $100$ \rgb images of $1$ minimally-clothed subject in a lab setting with \groundtruth \smplX meshes and \threeD scans.

\qheading{\agora~\cite{AGORA_2020}:}
\tdv{We evaluate whole-body \tdv{and body-only} accuracy on this, 
using its body-face-hands (BFH) subset.}
It has rendered \cite{unreal} photo-realistic images of \threeD human scans \cite{threedpeople,axyz,humanalloy,renderpeople} in scenes \cite{unrealMarketplace,hdrihaven}.
It has \smplx ground truth recovered from scans, images and semantic labels \cite{shape_under_cloth:CVPR17}. 

\qheading{\threeDPW~\cite{vonmarcard_eccv_2018_3dpw}:}
We evaluate main-body accuracy on this.
It captures $5$ subjects in indoor/outdoor videos with \smpl pseudo ground truth, recovered from images and IMUs. 

\qheading{\NOW~\cite{Sanyal2019_ringnet}:}
We use it to evaluate face/head-only accuracy. 
It contains \threeD head scans for $100$ subjects, and $2054$ images with various viewing angles and facial expressions.

\qheading{\freihand~\cite{Freihand2019}:}
We evaluate hand-only accuracy on this.
It has $37k$ hand/hand-object images of $32$ subjects, with \mano ground truth, recovered from multi-view images.

%% file: 04_experiments_02_metrics.tex
\subsection{Evaluation Metrics}

\qheading{Mesh alignment:}
Prior to computing a metric, we align estimated meshes to \groundtruth ones. 
The prefix  ``\paalignment'' denotes  Procrustes Alignment (solving for scale, rotation and translation), 
while       ``\tralignment'' denotes  translation alignment.
``\tralignment'' is stricter, as it does not factor out scale and rotation. 
\cameraReady{When reporting hand-/face-only metrics for the
full body, we align each part separately.}

\qheading{Mean Per-Joint Position Error (\mpjpe)}:
We report the mean Euclidean distance between the estimated and \groundtruth joints. 
For the body-only metric, we compute the $14$ LSP-common joints \cite{johnson2010clustered} as a common skeleton across different body models, 
using a linear joint regressor \cite{bogo2016keep,Lassner:UP:2017} on the estimated and \groundtruth vertices. 
This is a standard metric, but is too sparse; it cannot capture errors in full \threeD shape (\ie surface), or all limb rotation errors. 

\qheading{Vertex-to-Vertex (\VtoV):} 
For methods that infer meshes with the same topology as the \groundtruth ones, \eg \smplorsmplx estimations and \smplorsmplx ground truth, we compute the mean per-vertex error by taking into account \emph{all} vertices. 
This is not possible for methods with different topology, \eg~\smpl estimations for \smplX ground truth, and vice versa.
For such cases, we compute a \emph{main-body} variant of \VtoV,
\tdv{\ie without the hands and head},
as \smpl and \smplX share the same topology for the main body. %
\tdv{%
FB-\VtoV is the weighted sum 
of body (B), hand (LH, RH) and face (F) errors:
$\text{FB}= \text{B} + \frac{\text{LH} + \text{RH} + \text{F}}{3}$.
} \VtoV is stricter than \mpjpe; it also captures \threeD shape errors and unnatural limb rotations (for the same joint positions). 

\qheading{Point-to-Surface (\PtoS):} 
To compare \modelname with methods that use a different mesh topology to \smplorsmplx, \eg \mtc \cite{Xiang2019}, we measure the mean distance from \groundtruth vertices to the \emph{surface} of the estimated mesh.
\PtoS is stricter than \mpjpe; it captures errors in \threeD shape, but not unnatural limb rotations (for the same joint positions).

%% file: 04_experiments_03_eval_quant.tex
\subsection{Quantitative Evaluation}

\par\textbf{Whole-body.}
In \tab{tab:ehf_full} - \ref{tab:ehf_ablative} we report whole-body metrics (``All''), by taking into account the body, face and hands jointly.
We add body-only (``Body''), hand-only (``L/R hand''), and face-only (``Face'') variants for completeness.

\qheading{\ehf \cite{Pavlakos2019_smplifyx}:}
\Tab{tab:ehf_full} compares \modelname to three baseline sets: 
(1) the optimization-based 		\smplifyX \cite{Pavlakos2019_smplifyx} and \mtc \cite{Xiang2019} 				    that infer  \smplx and \adam, 
(2) the regression-based		 	\spin \cite{Kolotouros2019_spin} 												that infers \smpl, and
(3) the regression-based 		\expose \cite{Choutas2020_expose} and \frankmocap \cite{rong2020frankmocap} 	    that infer  \smplx. 
Note that \mtc does not estimate the face. 
\modelname outperforms optimization methods on most metrics, while being significantly faster.
Moreover, it is on par with regression methods, both in terms of error metrics and runtime, which drops to $0.08$ sec for known body-part crops.

\input{paper_FIG_07_AGORA_TEST}
\input{paper_TAB_03_experiments_EHF_ablative}

\qheading{\agora \cite{AGORA_2020}:} \tdv{%
\Fig{fig:agora_test} compares \modelname
to %
whole-body \cite{Pavlakos2019_smplifyx,rong2020frankmocap,Choutas2020_expose}
and 
body-only \cite{Kanazawa2018_hmr,Kolotouros2019_spin,pare,lin2021end-to-end,joo2020eft,romp}
regressors, for a varying occlusion degree. 
\modelname outperforms all methods, 
and is competitive on body-only metrics even to the occlusion-aware \pare \cite{pare}.}
\cameraReady{Note that \agora is much more complex and natural than \ehf, making the results more representative of real-world scenarios}.

\qheading{Ablation for moderators:}
\Tab{tab:ehf_ablative} compares
\modelname to naive whole-body regression (no body-part experts)
and the \cameraReady{``copy-paste''} fusion strategy.
\cameraReady{The latter copies pose parameters from the part experts (see \cite{Choutas2020_expose,rong2020frankmocap}),
as well as shape parameters from the face expert, 
to the whole body.} 

The naive version does not benefit from the expertise of the part experts.
``Copy-paste'' fusion can lead to erroneous hand/face orientation inference, since the respective experts lack global context.
Moreover, estimating whole-body shape from a face image is not always reliable, \eg when a person faces away from the camera (\fig{fig:experts_confidence}).
\modelname %
fuses ``global'' body and ``local'' part features with its moderators. 
\tdv{
In this way, 
it estimates more accurate \threeD bodies
and is more robust to challenging ambiguities (blur, occlusion)
than existing whole-body regressors,
especially on %
stricter metrics without Procrustes alignment.}\\
\qheading{Ablation for ``gendered'' shape loss on \threedpw \cite{vonmarcard_eccv_2018_3dpw}:}
By removing 
our ``gendered'' shape loss, the \paalignment-\VtoV error increases from $50.9$ to $51.7$ mm.
A qualitative ablation is shown in \fig{fig:gender_shape_prior}; learned implicit reasoning about gender gives more realistic body shapes.
\smplX's shared shape space for the whole body \highlight{lets parts 
contribute to 
the whole}.%
\par \textbf{Parts-only:}
For completeness, we use standard benchmarks for body-only, face-only, and hand-only evaluation.

\qheading{Body-only on \threedpw \cite{vonmarcard_eccv_2018_3dpw}:}
\Tab{tab:pose_results_3dpw} shows that \modelname performs on par \frankmocap \cite{rong2020frankmocap} and \expose \cite{Choutas2020_expose}
\cameraReady{and is worse than \spin \cite{Kolotouros2019_spin}},
for the \paalignment-\mpjpe metric,
but outperforms them all in the stricter \tralignment-\mpjpe (joints) and \VtoV (surface) metrics.

\qheading{Face-only on \now \cite{Sanyal2019_ringnet}:}
\Tab{tab:now} shows that \modelname outperforms not only the expressive whole-body method \expose \cite{Choutas2020_expose}, 
but also strong and dedicated face-only methods,
\tdv{except for the recent work of Feng \etal~\cite{DECA_2020}.}
\\
\qheading{Hand-only on \freihand \cite{Freihand2019}:}
\Tab{tab:freihand} shows that our hand expert performs on par with the whole-body %
\expose \cite{Choutas2020_expose}, \cameraReady{is a bit worse than} 
the hand-specific ``\mano CNN'' \cite{Freihand2019}, but %
outperforms the hand expert of Zhou \etal \cite{zhou2021pixieLike}.

\input{paper_TAB_04_experiments_POSE_3DPW}
\input{paper_TAB_05_NOW}
\input{paper_TAB_06_HAND_EVAL}

%% file: paper_FIG_07_AGORA_TEST.tex
\begin{figure*}
    \centering
    \includegraphics[trim=000mm 000mm 000mm 000mm, clip=true, height=0.23 \linewidth]{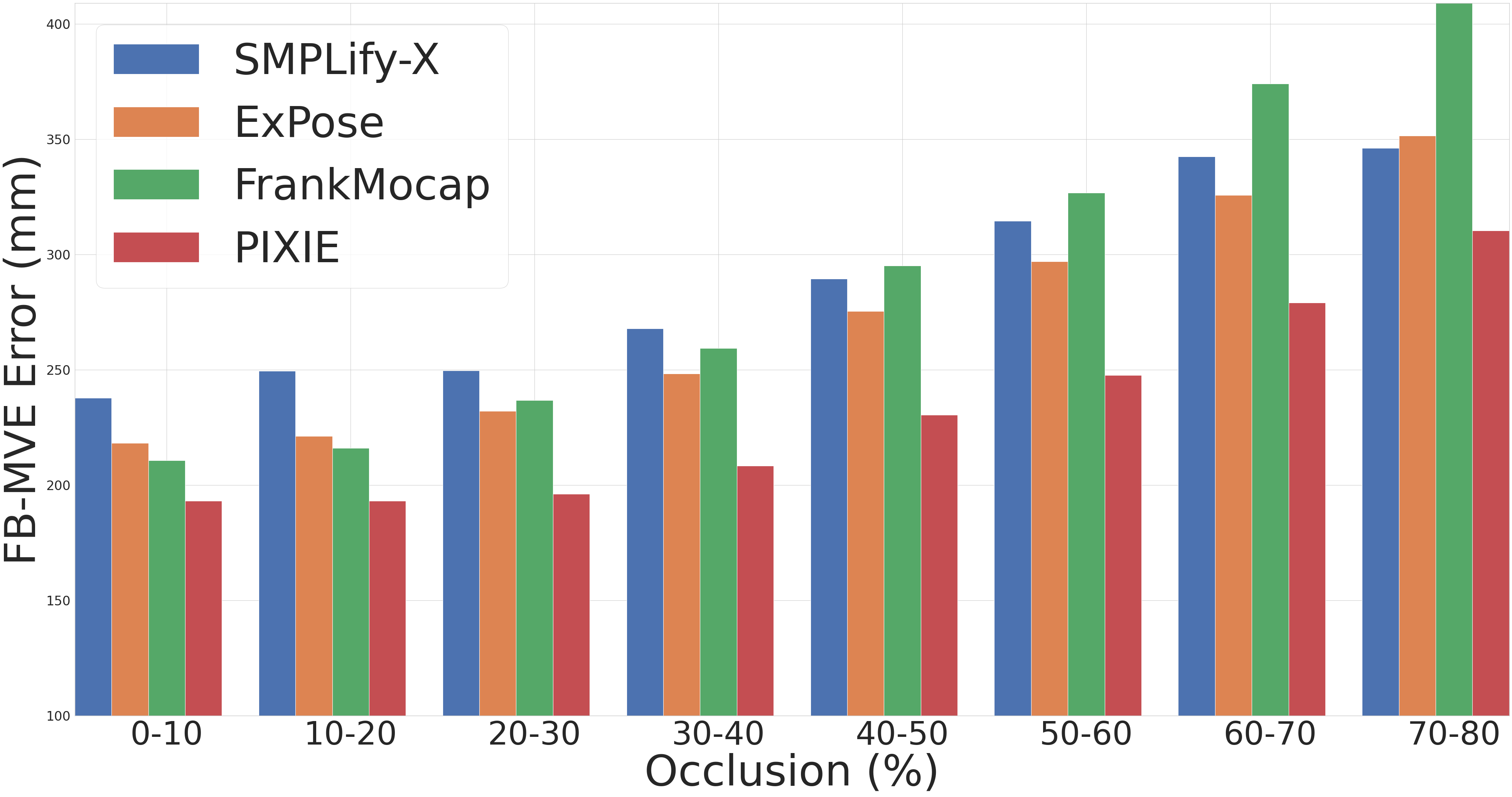}
    \quad
    \includegraphics[trim=000mm 000mm 000mm 000mm, clip=true, width=0.51 \linewidth]{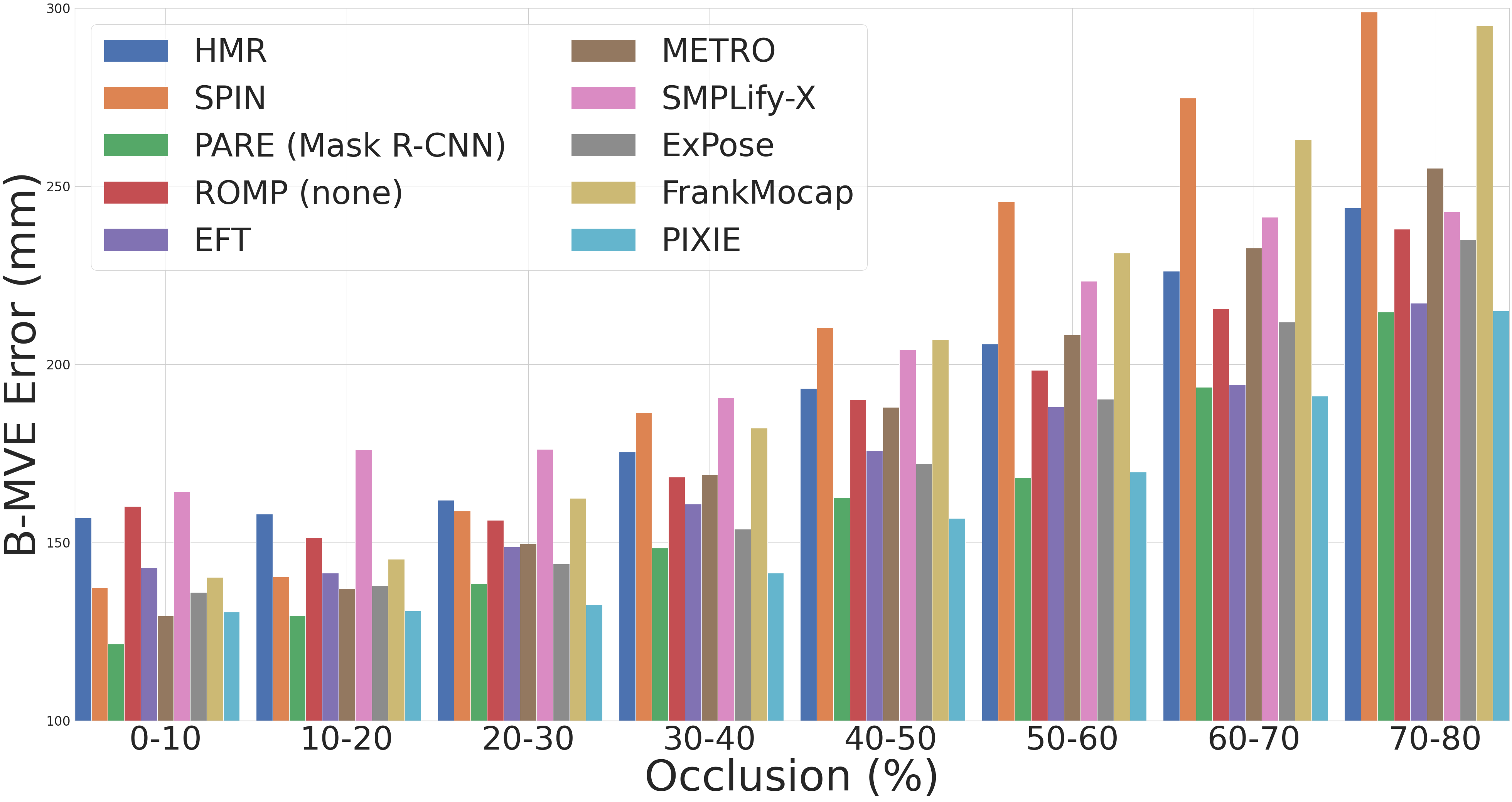}
    \vspace{-0.7em}
    \caption{%
    \tdv{
        Comparison against state-of-the-art full-body (left) and body-only (right) methods on \agora~\cite{AGORA_2020}, using 
        the vertex-to-vertex (\VtoV) metric
        (mm) for varying percentages of occlusion. 
        \cameraReady{Unless otherwise noted (in
        parens),
        we use OpenPose to extract person bounding boxes.}
        \modelname %
        outperforms 
        existing methods, including 
        the occlusion-aware \pare \cite{pare}. %
        }
    }
    \vspace{-1.5em}
    \label{fig:agora_test}
\end{figure*}

%% file: paper_TAB_03_experiments_EHF_ablative.tex
\begin{table}
    \centering
    \scriptsize
    \vspace{+0.8em}
    \begin{tabular}{l|c|c|c|c}
        \toprule
        \multirow{2}{*}{Method}
        &  \multicolumn{2}{c}{\paalignment-\VtoV (mm) $\downarrow$}
        &  \multicolumn{2}{|c}{\tralignment-\VtoV (mm) $\downarrow$}
        \\
        \cline{2-5}
        	{}
        	&  All			&	Body
        	&  All			&	Body
        	\\
        \midrule
        Naive Body
        & 	59.7			&	54.3
        & 70.5 & 83.4 \\
        ``Copy-paste''		&  60.3			&	55.5
        & 72.9 & 82.4 \\
        \modelname (ours)
        & \textbf{55.0}	& \textbf{53.0}
        & \textbf{67.6} & \textbf{75.8} \\
        \bottomrule
    \end{tabular}
    \vspace{-0.5em}
    \caption{
        Ablation for our moderator on \ehf \cite{Pavlakos2019_smplifyx}.
        ``Naive body'' denotes a single regressor for the whole body, and
        ``Copy-Paste'' denotes a naive integration of the independent expert estimations on the inferred body.
    }
    \label{tab:ehf_ablative}
\end{table}

%% file: paper_TAB_04_experiments_POSE_3DPW.tex
\begin{table}
    \scriptsize
\resizebox{1.00\linewidth}{!}{
    \centering
    \begin{tabular}{l|l|c|c|c}
        \toprule
        \multirow{2}{*}{Method} & Body  & \paalignment-\mpjpe & \tralignment-\mpjpe & Body \paalignment-\VtoV \\
        {}                      & model & (mm) $\downarrow$   & (mm) $\downarrow$   & (mm) $\downarrow$
        \\
        \toprule
        \hmr       \cite{Kanazawa2018_hmr}    & \smpl		& 81.3              & 130.0             & 65.2
        \\
        \spin      \cite{Kolotouros2019_spin} & \smpl		& \textbf{59.2}     & 96.9              & 53.0
        \\
        FrankMocap \cite{rong2020frankmocap}  & \smplx		& 61.9              & 96.7              & 55.1
        \\
        ExPose     \cite{Choutas2020_expose}  & \smplx		& 60.7              & 93.4              & 55.6
        \\
        \modelname (ours)                     & \smplx		& 61.3              & \textbf{91.0}     & \textbf{50.9}
        \\
        \bottomrule
    \end{tabular}
    }
    \caption{
        Evaluation on \threedpw \cite{vonmarcard_eccv_2018_3dpw}.
        \modelname is the best for the stricter \tralignment-\mpjpe (joints) and \VtoV (surface) metrics.
    }
    \label{tab:pose_results_3dpw}
\end{table}

%% file: paper_TAB_05_NOW.tex
\begin{table}
    \centering
    \scriptsize
    \begin{tabular}{l|c|c|c}
    \toprule
    \multirow{2}{*}{Method} &
    \multicolumn{3}{c}{\paalignment-\PtoS for face/head    (mm) $\downarrow$}
     \\
    \cline{2-4}
    {} 
    & Median (mm) $\downarrow$
    & Mean (mm)   $\downarrow$
    & Std (mm)    $\downarrow$
    \\
    \toprule
    3DMM-CNN   \cite{AnhTran2017}        & 1.84   & 2.33  & 2.05    \\ 
    PRNet      \cite{Feng2018}           & 1.50   & 1.98  & 1.88    \\ 
    Deng \etal~\cite{Deng2019}           & 1.23   & 1.54  & 1.29    \\ 
    \ringnet   \cite{Sanyal2019_ringnet} & 1.21   & 1.54  & 1.31    \\ 
    3DDFA-V2   \cite{guo2020towards}     & 1.23   & 1.57  & 1.39    \\ 
	\deca      \cite{DECA_2020}          & \textbf{1.09}  & \textbf{1.38} & \textbf{1.18}  \\
    \midrule
    \expose \cite{Choutas2020_expose}    & 1.26   & 1.57  & 1.32  \\ 
    \modelname (ours)                    & \textit{1.18}   &
    \textit{1.49}  & \textit{1.25}  \\ 
    \bottomrule
    \end{tabular}
    \vspace{-0.5em}
    \caption{
        Evaluation on \now~\cite{Sanyal2019_ringnet}.
        \modelname is better than the whole-body \expose, 
        \cameraReady{it outperforms many strong face-specific methods, 
        and is a bit worse than \deca \cite{DECA_2020}}. 
    }
    \label{tab:now}
    \vspace{-0.5em}
\end{table}

%% file: paper_TAB_06_HAND_EVAL.tex
\begin{table}
    \scriptsize
    \centering
    \begin{tabular}{l|c|c|c|c}
        \toprule
        \multirow{2}{*}{Method} 
        & {\paalignment-\mpjpe}
        & {\paalignment-\VtoV}
        & {\paalignment-F@}               & {\paalignment-F@}               \\
        & {(mm) $\downarrow$}    & {(mm) $\downarrow$}    & {5mm $\uparrow$}   & {15mm $\uparrow$}  \\
        \midrule
        ``\mano CNN'' \cite{Freihand2019} & \textbf{11.0} & \textbf{10.9} & \textbf{0.516} & \textbf{0.934}  \\
        \midrule
        \expose   \cite{Choutas2020_expose} hand expert 	  & 12.2 & \textit{11.8} & \textit{0.484} & 0.918  \\
        Zhou \etal \cite{zhou2021pixieLike} & 15.7 & - & - & - \\
        \modelname                          hand expert & \textit{12.0} & 12.1 & 0.468 & \textit{0.919}  \\
        \bottomrule
    \end{tabular}
    \vspace{-0.5em}
    \caption{
        Evaluation on \freihand \cite{Freihand2019}.
        \cameraReady{\modelname's hand expert is on par with the hand expert of \expose,
        but clearly outperforms the more related Zhou \etal \cite{zhou2021pixieLike}
        that also uses hand-body feature fusion.}
	}
    \label{tab:freihand}
    \vspace{-0.5em}
\end{table}

%% file: 04_experiments_04_eval_qual.tex
\subsection{Qualitative Evaluation}

\input{paper_FIG_06_GENDER_ABLATION}

\input{paper_FIG_04_PIXIE_vs_FrankMocap}
\input{paper_FIG_05_PIXIE_vs_CVPR21}

\Fig{fig:qual_comparison_frank} compares \modelname with \frankmocap \cite{rong2020frankmocap} and \expose \cite{Choutas2020_expose}, which also regresses \smplx.
Both baselines fail when the hand expert faces ambiguities (row $2$); \modelname gains robustness by using the full-body context. 
Both baselines give body shapes that \cameraReady{look average (rows $1$, $4$)} or have the \cameraReady{wrong gender (rows $2$, $3$)}; \modelname gives the most realistic shapes due to its ``gendered'' shape loss.
\frankmocap fails for \cameraReady{strong occlusions (rows $1$, $3$)}.
Lastly,   \expose struggles with accurate facial expressions, and \frankmocap with \cameraReady{head rotations (rows $1$, $3$)};
\modelname outperforms both with its strong face/head expert and predicts a more realistic face.

\Fig{fig:qual_comparison_zhou} compares \modelname with Zhou \etal \cite{zhou2021pixieLike}, recent work that also estimates a textured face.
\modelname gives more accurate poses (see how hands and faces align to the image), 
as it fuses both face-body and hand-body expert features, %
weighted by their confidence.
\modelname also gives more realistic body shapes, both due to its gendered shape loss and due to 
\highlight{part experts contributing to whole-body shape},
using \smplX's shared body, hand and face shape space.

%% file: paper_FIG_06_GENDER_ABLATION.tex
\begin{figure}
    \centering
    \includegraphics[trim=025mm 000mm 010mm 000mm, clip=true, width=1.00 \columnwidth]{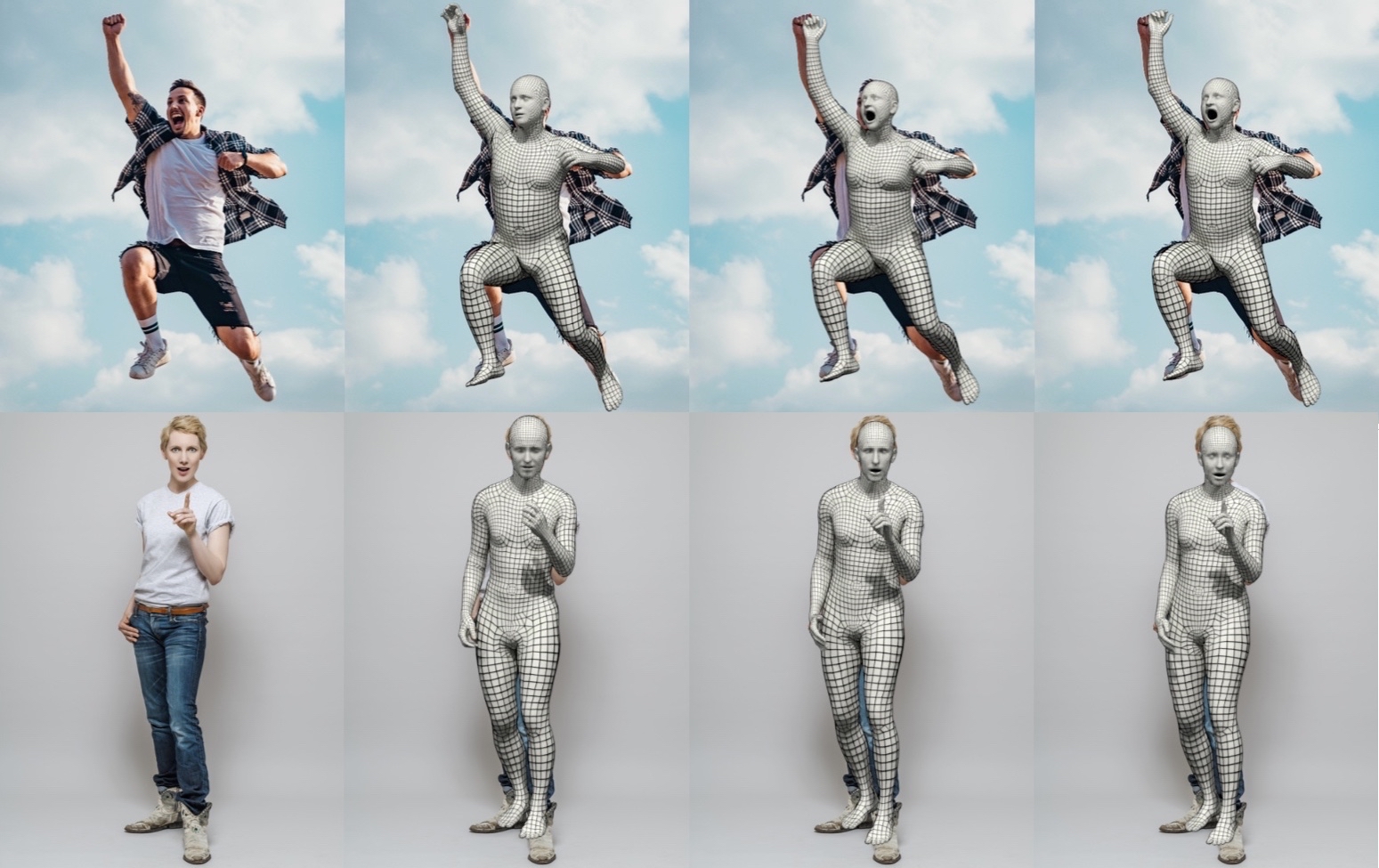}
    \vspace{-1.5em}
    \caption{
    	Ablation for the ``gendered'' shape loss %
    	and the shared shape space \tdv{(body/head)}. %
        From left to right:
        (1)     RGB Image,
        (2)     \tdv{shape prediction only from the body image}, and 
                \modelname without (3) and with (4) the ``gendered'' shape \cameraReady{loss}. %
        \tdv{We always use the gender-neutral \smplx model.}
    }
    \vspace{-0.5em}
    \label{fig:gender_shape_prior}
\end{figure}

%% file: paper_FIG_04_PIXIE_vs_FrankMocap.tex
\begin{figure}%
    \centering
    
    \includegraphics[trim=000mm 000mm 000mm 000mm, clip=true,width=1.00\linewidth]{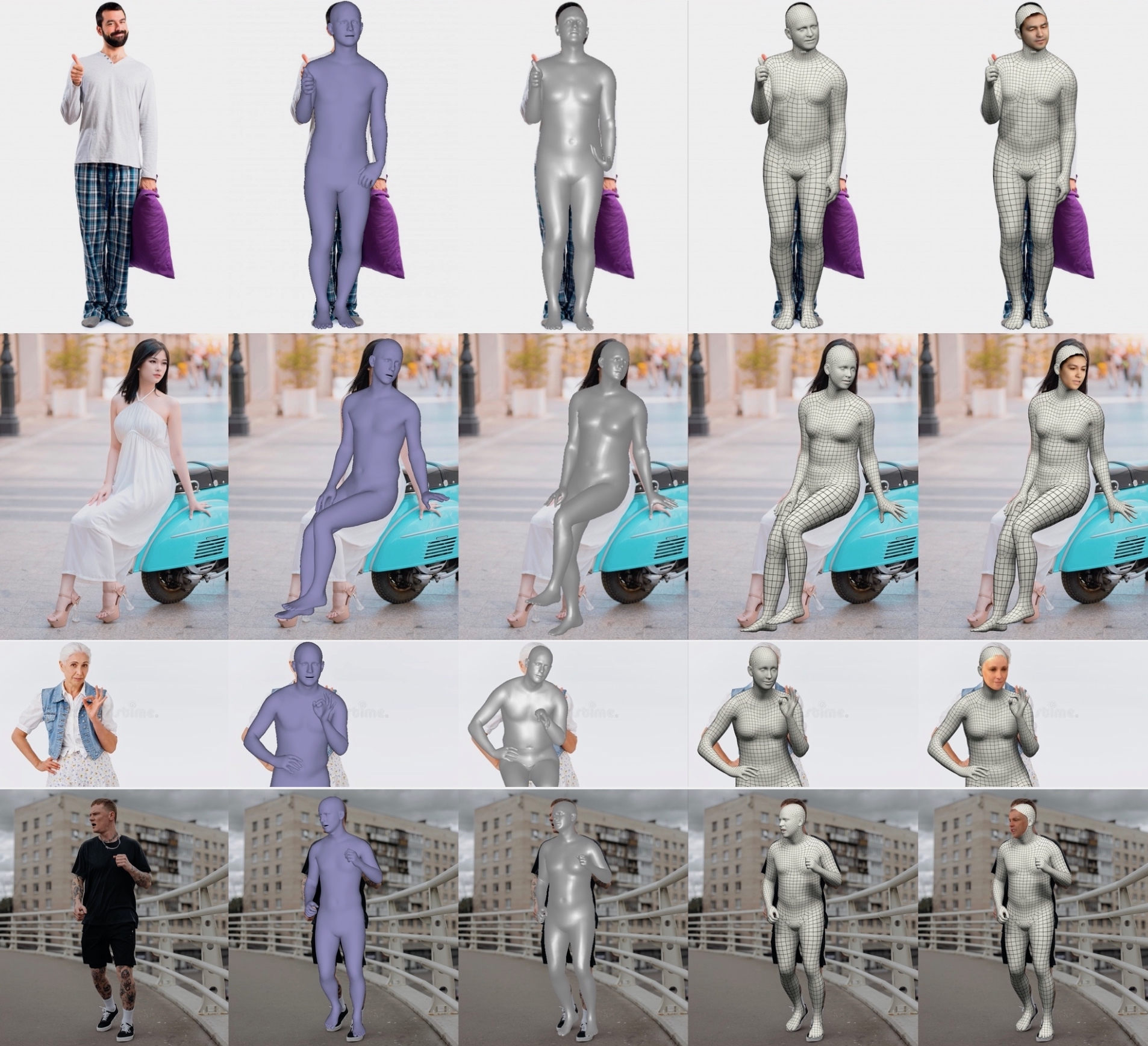}

    \vspace{-0.5em}
    \caption{%
    \tdv{%
        Qualitative comparison. 
        From left to right:
        (1) \rgb Image,
        (2) \expose \cite{Choutas2020_expose},
        (3) \frankmocap \cite{rong2020frankmocap},
        (4) \tdv{\modelname,}
        (5) \modelname with predicted albedo and lighting.
        }
    }
    \vspace{-0.5em}
    \label{fig:qual_comparison_frank}
\end{figure}

%% file: paper_FIG_05_PIXIE_vs_CVPR21.tex
\begin{figure}
    \centering%
    \includegraphics[trim=000mm 002mm 000mm 000mm, clip=true, width=0.99\linewidth]{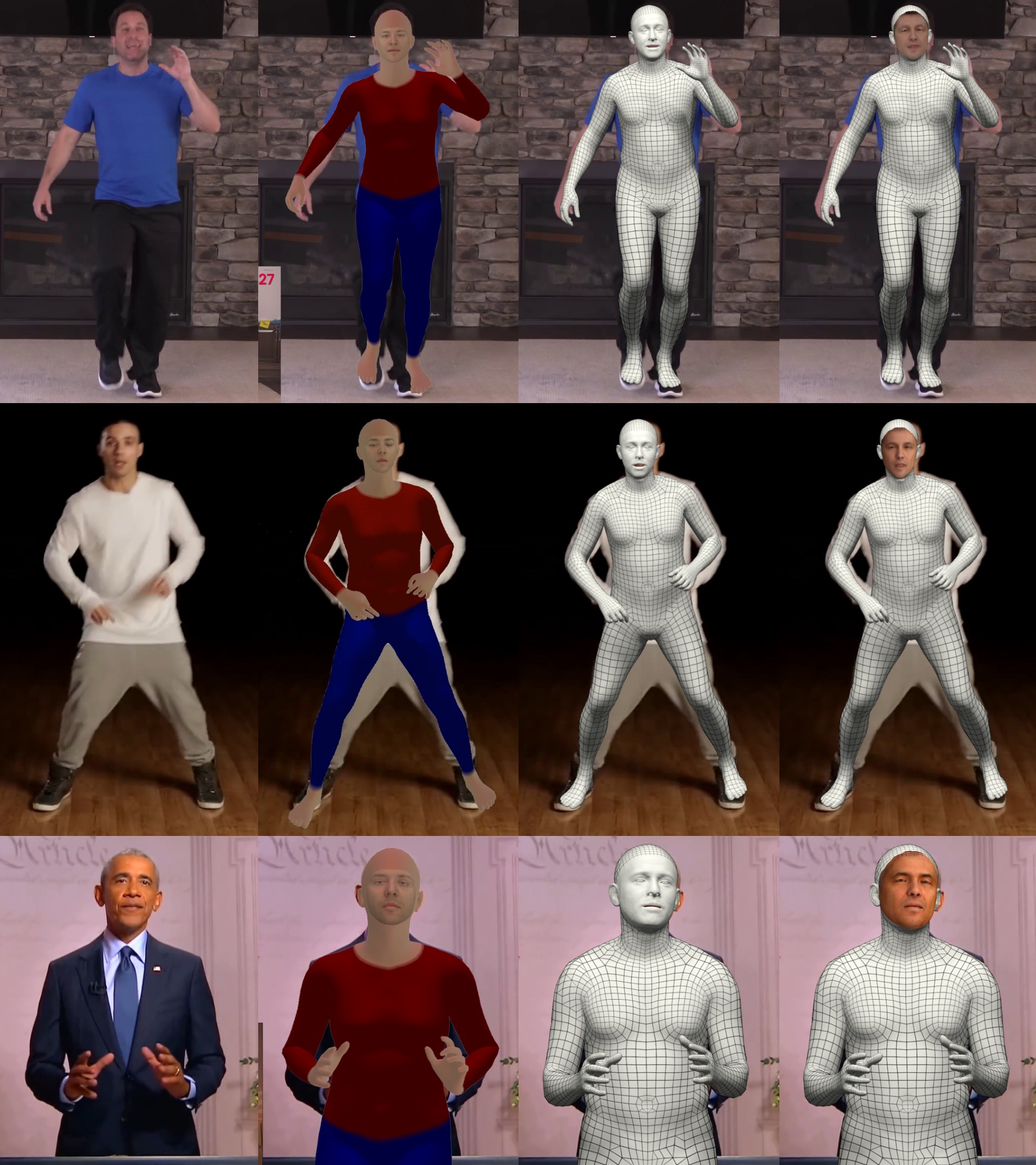}
    \vspace{-0.5em}
    \caption{
        Comparison with Zhou \etal \cite{zhou2021pixieLike}. 
        From left to right:
        (1) \rgb image,
        (2) Zhou \etal,
        (3) \modelname with inferred facial details and 
        (4)                 inferred albedo and lighting. 
        Note that Zhou \etal use tight face crops through Dlib \cite{king2009dlib} to improve performance; \modelname needs no tight face crops. 
    }
    \label{fig:qual_comparison_zhou}
\end{figure}

%% file: 04_experiments_05_future.tex
\tdv{
\qheading{Future work:}
Mesh-to-image misalignment is a common limitation of regressors that pool ``global'' features from the image, losing local information.
This could be tackled with ``pixel-aligned'' features \cite{Guler_2019_CVPR,pare,Saito_2019_ICCV,zhang2020learning}.
Moreover, \smplX models bodies without clothing; adding clothing models \cite{corona2021smplicit,ma2020cape} is a challenging but promising avenue.
Furthermore, due to the formulation of the photometric term the model prefers
to explain image evidence using lighting, rather than albedo,
which leads to wrong skin tone predictions.
Future work could further improve cases with self-contact \cite{mueller2021tuch,selfcontact_aaai_2201}, or other extreme ambiguities.
}

%% file: 05_conclusion.tex
\section{Conclusion}

We present \modelname, a novel expressive whole-body reconstruction method that recovers an animatable \threeD avatar with a detailed face from a single \rgb image.
\modelname uses body-driven attention to leverage dedicated body, head and face experts.
It learns a novel moderator that reasons about the confidence of each expert, to fuse their features according to confidence, and exploit their complementary strengths.
It uses the best practices from the face community for accurate faces with realistic albedo and geometric details.
The face expert can contribute to more realistic whole-body shapes, by using a shared face-body shape space.
To further improve shape, \modelname uses implicit reasoning about gender, to encourage likely ``gendered'' body shapes.
Qualitative results show natural and expressive humans, with improved body shape, well articulated hands, and realistic faces, comparable to the best face-only methods.
We believe that \modelname will be useful for many applications that need expressive human understanding from images.

%% file: 06_disclosure_acks.tex
\\
{%
\noindent
\small
\textbf{Acknowledgments:}{We thank Victoria Fernández Abrevaya, Yinghao Huang, Yuliang Xiu, Radek Danecek for discussions and Priyanka Patel for \agora experiments.
This work was partially supported by the Max Planck ETH Center for Learning Systems.}
\\
\href{https://files.is.tue.mpg.de/black/CoI/3DV2021.txt}{
\textbf{\color{black}Disclosure: }https://files.is.tue.mpg.de/black/CoI/3DV2021.txt
}
}

%% file: sup_mat.tex
\renewcommand{\thefigure}{A.\arabic{figure}}
\setcounter{figure}{0}
\renewcommand{\thetable}{A.\arabic{table}}
\setcounter{table}{0}

\newpage
\begin{appendices}
    \input{supmat_01_training_details}
    \input{supmat_02_evaluation}
\end{appendices}

%% file: supmat_01_training_details.tex
\section{Implementation Details}

\qheading{Data augmentation:} 
For training data, we use image crops around the body, face and hands. 
We augment our training image crops, following mainly \cite{Choutas2020_expose}, as described below. 
First, we use standard techniques, namely random horizontal flipping, random image rotations, color noise addition and random translation of the crop's center. 
However this is not enough, as there is a significant domain gap between face-only and hand-only datasets, and the respective image crops extracted from full-body images; the former have significantly higher resolution. 
To account for this, we also randomly down-sample and up-sample the head and hand image crops, to simulate various lower resolutions. 
Finally, inspired by \cite{rong2020frankmocap}, we add synthetic motion blur to face and hand crops, to simulate the motion blur that is common in full-body images.

\qheading{Training details:}
We use \pytorch \cite{pytorch} to implement our pipeline. 
We follow a three-step training procedure: 
(1) 	We pre-train the model with body-only, face-only and hand-only datasets; for each dataset we train only the respective parameters. 
	Since these datasets are captured independently, there is no body image that corresponds to a face-only or hand-only image. 
	Consequently, for this step we cannot apply feature fusion, and 
	body-part features 
	go directly to the respective regressor(s) (bypassing the moderators), 
	to estimate the respective body-part parameters. 
	Similar to existing work, we train only a right hand regressor; for images of a left hand, we flip the image horizontally to use the right hand regressor, and mirror the predictions to get a left hand. 
(2)	Then, using the same data, we freeze the feature encoders and proceed with training the regressors and extractors
(see Fig. $3$ of the paper the linear layers
$\extractor^{\hand}$ and $\extractor^{\face}$ between the
body encoder $\encoder_{\body}$ and moderators
$\moderator_{\hand}$ and $\moderator_{\face}$ respectively). 
This step encourages features 
$F_{\body}^{\hand}$ and $F_{\body}^{\face}$ from body images 
to be in the same space as features $F_{\hand}$   and $F_{\face}$ 
from part-only images, so that 
regressors 
$\regressorFaceFused$ and 
$\regressorHandFused$
work for both feature types. 
(3)	Finally, we train the full network, including the moderators
$\moderator_{\hand}$ and $\moderator_{\face}$, 
but this time using training images with full \smplX ground truth,
to extract part crops from full-body images as well. 
However, there are	two problems. 
First,  for these images there is no skin mask available, consequently we remove the loss for body shape $\shape$ and do not apply a photometric and identity loss on head crops.
Second, localizing the hands with body-driven attention is much harder compared to the head, due to the longer kinematic chain,
consequently we freeze the hand regressor $\mathcal{R}_{\hand}$
to avoid fine-tuning it with invalid inputs. 

All parameters are optimized using \adam \cite{adam} with a learning rate of $0.0001$. 
For training the body, hand and face sub-networks, we use a batch size of $16$ , $16$, and $8$, respectively. 
\cameraReady{The moderator is a fully connected network with the following structure: FC (2048, 1024), ReLU, FC (1024, 1).
}
All input images are resized to $224 \times 224$ pixels before feeding them to our network. 
\cameraReady{During inference, we extract the hand/face crops using the
hand and face locations from $R_b$'s output. 
Hand and face cameras are ignored when estimating full body pose.}

\qheading{Global to relative pose:}
The regressors 
$\regressorFaceFused$ and 
$\regressorHandFused$
estimate the absolute head and wrist orientation 
$\pose_\text{g}$, 
\ie~irrespective of the (parent) main body's pose. 
However, to ``apply'' these 
$\pose_\text{g}$ 
estimates on a \smplX body that is already
posed by $\regressor_{\body}$ with $\pose_{\body}$
(up to the wrist and neck, excluding them),
we need to express them relative to their parent in the kinematic skeleton: 
\begin{align}
	\pose_\text{relative} 	&= \boldsymbol{\Gamma}(\pose_\text{g}, \pose_{\body}),
\end{align}
where $\boldsymbol{\Gamma}$ is the chain transformation function according to \smplX's kinematic skeleton hierarchy.

%% file: supmat_02_evaluation.tex
\section{Evaluation}			
\label{sec:eval}

\subsection{Body-face correlations discussion}
PIXIE gives more realistic body shapes, not only due to its gendered shape loss, but also thanks to the shared body, hand
and face shape space of SMPL-X. This allows PIXIE’s face
expert to – uniquely – contribute to whole-body shape.
To verify this, we apply our face expert on face-only images and get the whole-body shapes of Fig. \ref{fig:shape_from_head}. These are not
only correctly “gendered”, but also have a plausible BMI.
For the sumo wrestler in Fig.~\ref{fig:shape_from_head}, 
\modelname predicts a body with higher \bmi (26.9) than the mean shape (26.1).
PIXIE is the only 3D whole-body estimation method that
explores such face-body shape correlations explicitly.
We believe that this is a useful insight and points the
community towards a new direction.

\subsection{Qualitative Evaluation}
\label{subsec:qual_eval}

\qheading{Comparison with \mtc}: 
In Fig. \ref{fig:mtc} we compare \modelname with \mtc \cite{Xiang2019}.
\modelname is two orders of magnitude faster and
predicts more accurate \threeD body shapes.
However, when \twoD joint estimations are accurate, optimization-based methods, such as \mtc \cite{Xiang2019} and \smplifyx \cite{Pavlakos2019_smplifyx}, tend to estimate bodies that are better aligned with the image.

\qheading{Expressive body reconstruction}: We compare our method, \modelname, with other state-of-the-art expressive body reconstruction methods in Fig. \ref{fig:qual_color}. 

\tdv{
PIXIE is more robust to challenging ambiguities (blur, occlusion) than existing whole-body regressors \cite{Choutas2020_expose,rong2020frankmocap}, since its moderators fuses “global” body and “local” part. 

}

\resubsupmat{

\qheading{Qualitative results}: Finally, in Fig. \ref{fig:pixie_only1},  \ref{fig:pixie_only2}
and \ref{fig:pixie_only3}
we provide more standalone \modelname
results. Overall, \modelname produces visually
plausible body shapes with detailed facial expressions.
}

\resubsupmat{
\qheading{Failure cases}: Although the gender prior loss and the shared whole-body shape space 
result in better \threeD shape predictions, they are not sufficient
for perfectly estimating full-body \threeD shape. 
Furthermore, the employed photometric term often causes the model
to prefer to explain image evidence using lighting, rather than albedo,
which leads to incorrect skin tone predictions.
These points highlight important directions for improving \modelname. 
Representative failure cases can be seen in Fig. \ref{fig:failures}.
}
\input{paper_FIG_07_Shape_from_head}
\input{supmat_FIG_01_MTC}
\input{supmat_FIG_02_MoreQUAL_color}
\input{supmat_FIG_02_PIXIE_only}
\input{supmat_FIG_03_FAIL}

%% file: paper_FIG_07_Shape_from_head.tex
\begin{figure}
\centering
\includegraphics[trim=000mm 000mm 000mm 000mm, clip=true, width=1.00\linewidth]{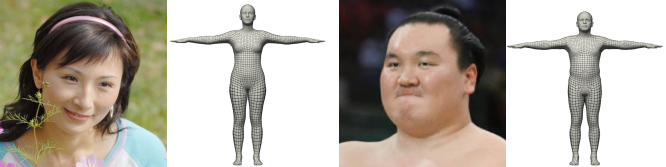}
\caption{
    Whole-body shape estimation from \emph{only} our face expert, using \smplX's joint shape space for all body parts.
}
\label{fig:shape_from_head}
\end{figure}

%% file: supmat_FIG_01_MTC.tex
\begin{figure*}[t!]
    \centering
    \includegraphics[width=1.000\textwidth]{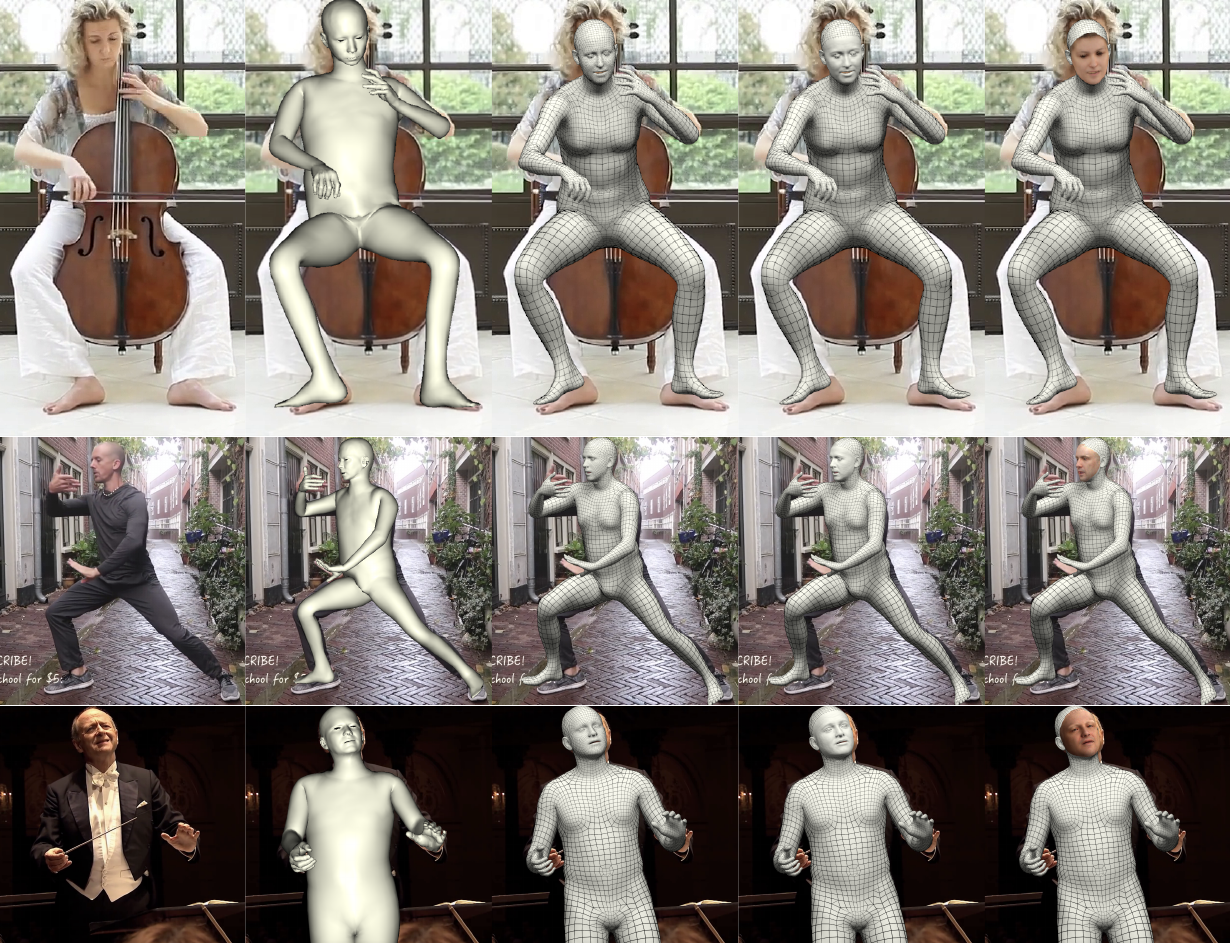}

   \caption{
   Qualitative \modelname results and comparison to \mtc \cite{Xiang2019}. 
   From left to right: 
   (1) \rgb image,  
   (2) \mtc \cite{Xiang2019}, 
   (3) \modelname, 
   (4) \modelname with facial geometric details, 
   (5) \modelname with estimated face albedo and lighting. 
   Overall, \modelname produces more visually plausible body shapes and more detailed facial expressions.
   }
   \label{fig:mtc}
\end{figure*}

%% file: supmat_FIG_02_MoreQUAL_color.tex
 \begin{figure*}
 	\centering
    \includegraphics[width=1.00\textwidth]{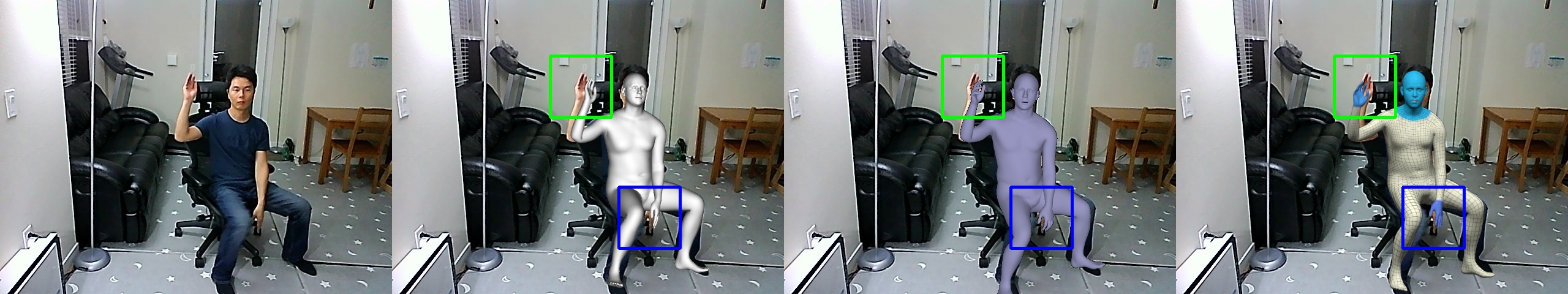}
    \includegraphics[width=1.00\textwidth]{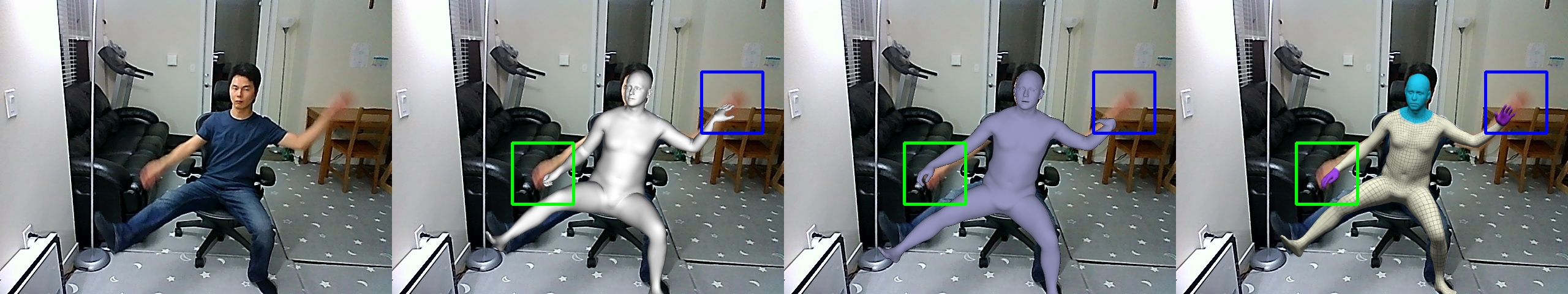}
    \includegraphics[width=1.00\textwidth]{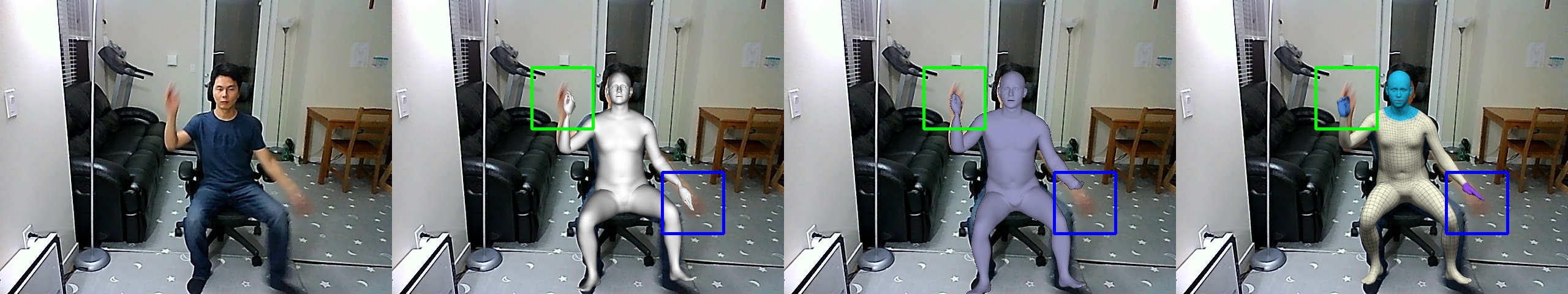}
    \includegraphics[width=1.00\textwidth]{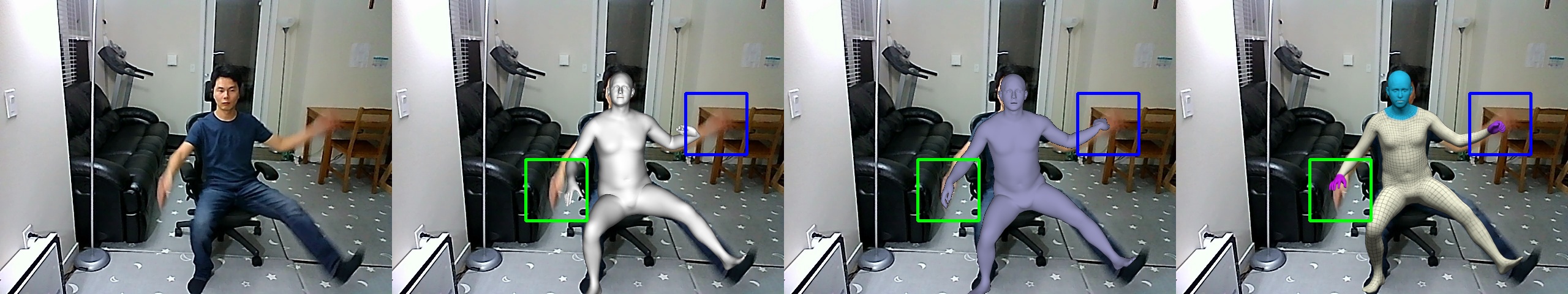}

     \caption{
     \tdv{
 				Qualitative \modelname results and comparison to \expose \cite{Choutas2020_expose} and \frankmocap \cite{rong2020frankmocap}. 
 				From left to right:
 				(1) \rgb images from \href{https://github.com/facebookresearch/frankmocap}{video},
 				(2) \frankmocap \cite{rong2020frankmocap},
 				(3) \expose \cite{Choutas2020_expose},
 				(4) \modelname \threeD body predictions with color-coded part-expert confidence. Moderator predicts the confidence of body/face/hand experts, reder means higher confidence in the body expert rather than the results from face/hand experts. 
 			Thanks to the moderators, PIXIE is more robust to low quality part images. For example, when the hand is blurry, \modelname still predicts a plausible wrist pose, instead of an unnatural twist.
 	}}
 	\label{fig:qual_color}
 \end{figure*}

%% file: supmat_FIG_02_PIXIE_only.tex
\renewcommand{\graphsize}{0.24\linewidth}

\begin{figure*}[t!]
    \centering
    \includegraphics[height=0.94\textheight]{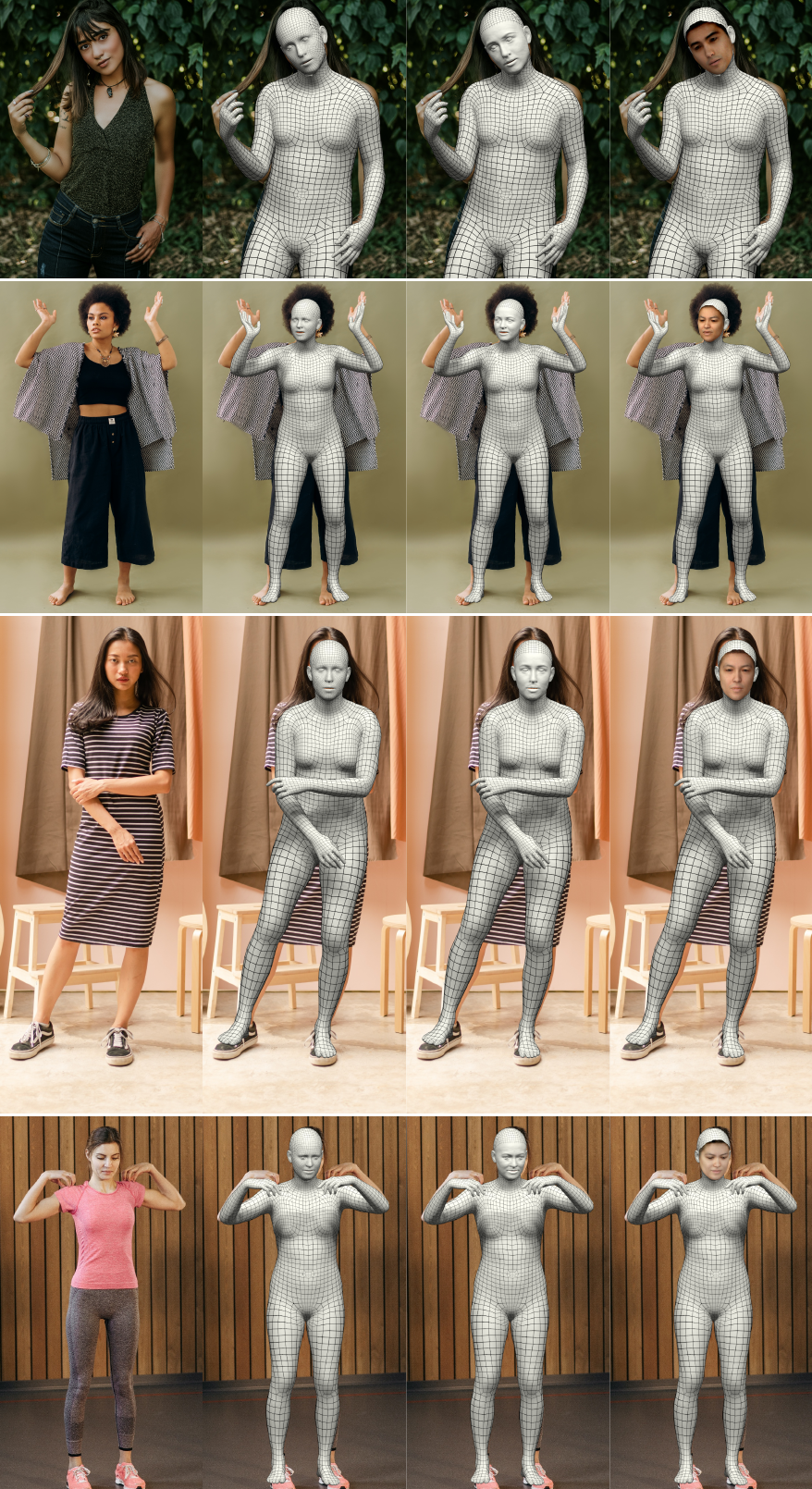}
   \caption{
   Qualitative \modelname results. 
   From left to right: 
   (1) \rgb image,  
   (2) \modelname, 
   (3) \modelname with facial geometric details, 
   (4) \modelname with estimated face albedo and lighting. 
   }
   \label{fig:pixie_only1}
\end{figure*}%

\begin{figure*}[t!]
    \centering
    \includegraphics[height=0.94\textheight]{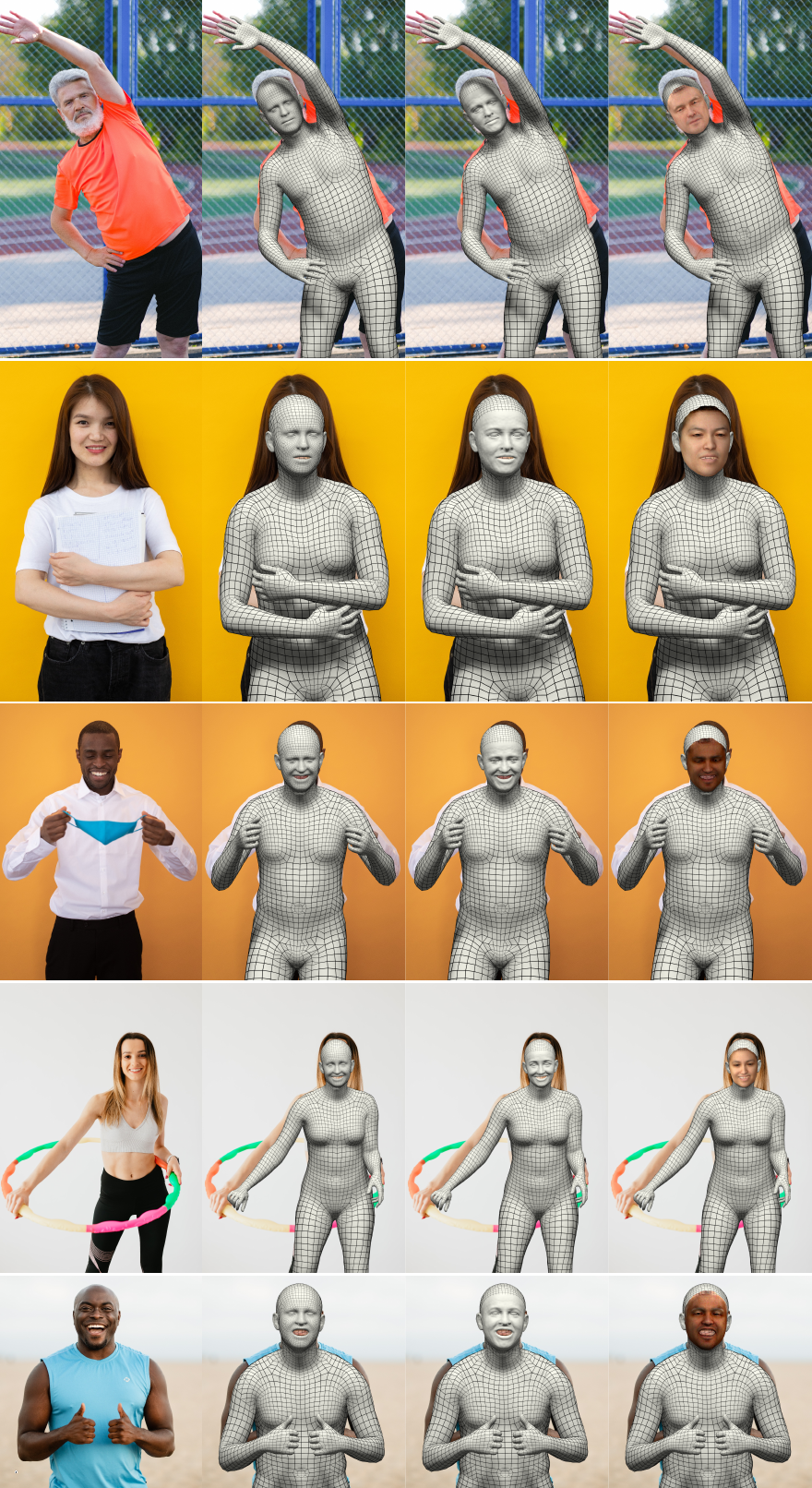}
   \caption{
   Qualitative \modelname results. 
   From left to right: 
   (1) \rgb image,  
   (2) \modelname, 
   (3) \modelname with facial geometric details, 
   (4) \modelname with estimated face albedo and lighting. 
   }
   \label{fig:pixie_only2}
\end{figure*}%

\begin{figure*}[t!]
    \centering
    \includegraphics[height=0.92\textheight]{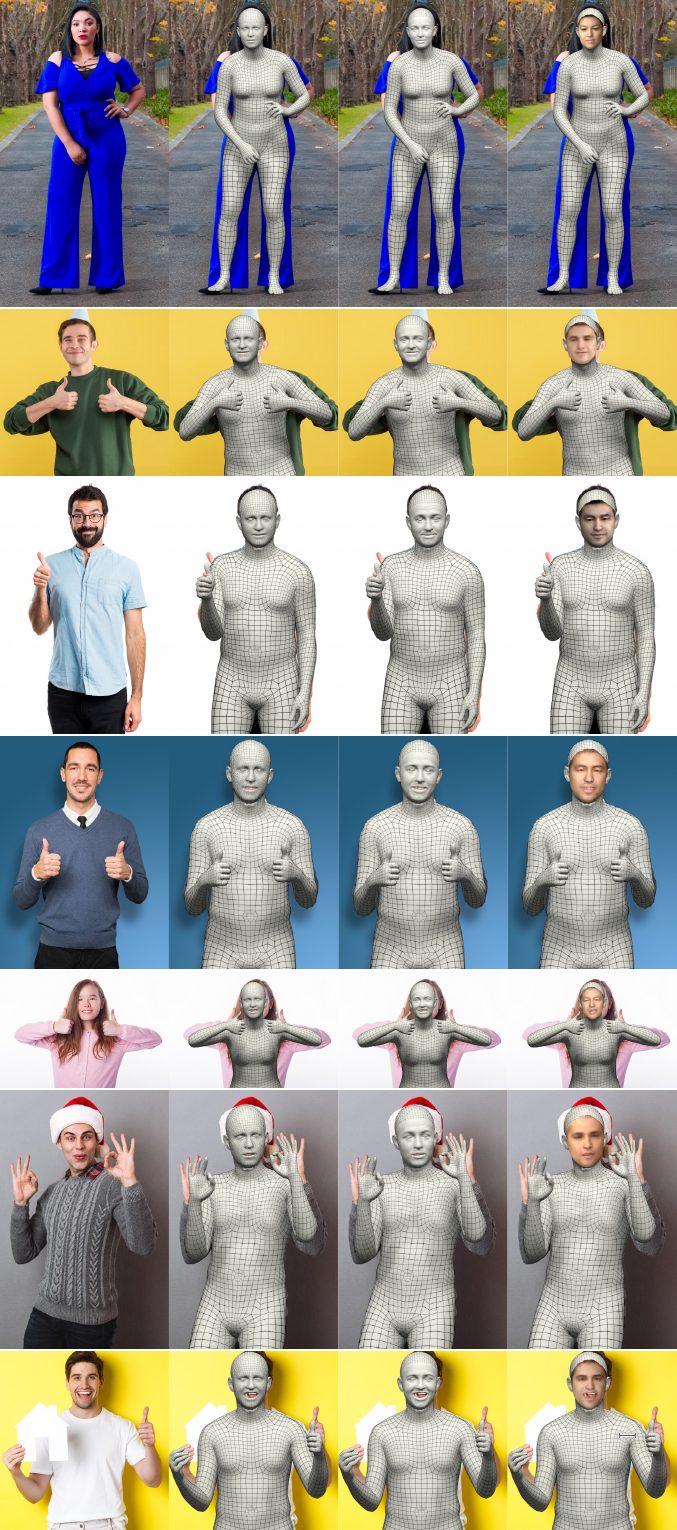}

   \caption{
   Qualitative \modelname results. 
   From left to right: 
   (1) \rgb image,  
   (2) \modelname, 
   (3) \modelname with facial geometric details, 
   (4) \modelname with estimated face albedo and lighting. 
   }
   \label{fig:pixie_only3}
\end{figure*}%

%% file: supmat_FIG_03_FAIL.tex
\begin{figure*}[t!]
	\centering
	
	\includegraphics[width=1.0\linewidth]{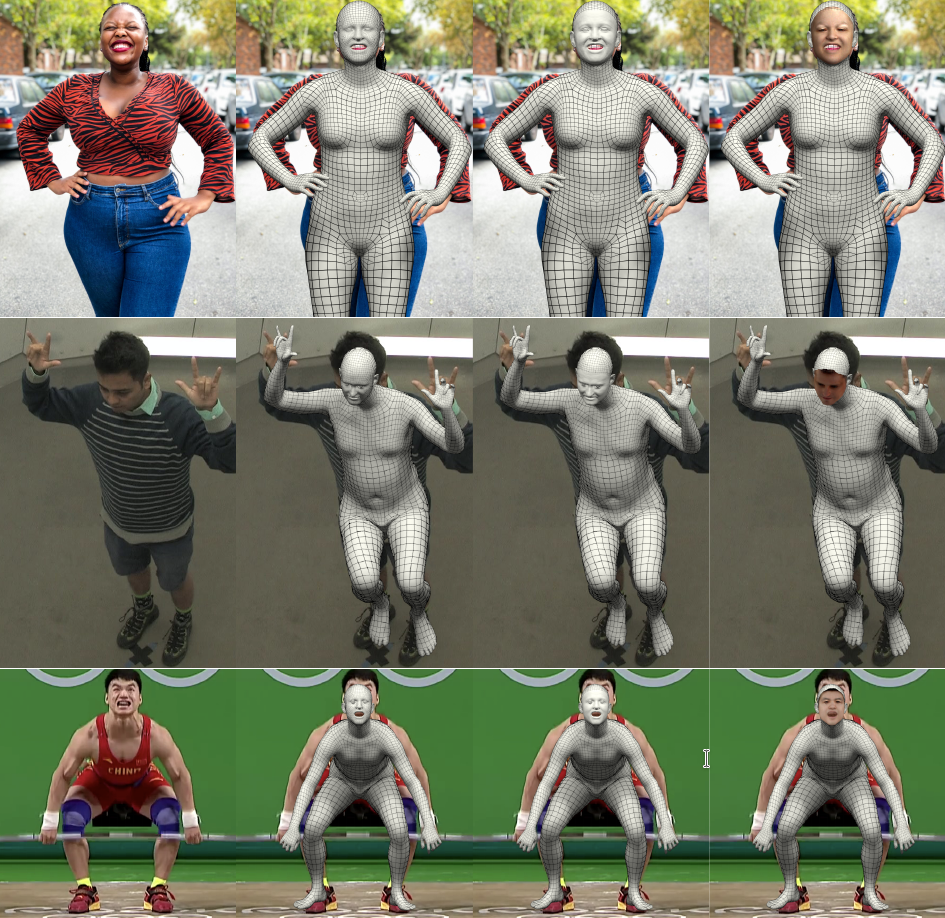}
	%
 %
 %
%
	
	\caption{
	Failure cases for \modelname.
	In these examples, the implicit reasoning about gender
	and the face shape information are not enough to correctly infer the body shape.
	\resubsupmat{
	Furthermore, due to the formulation of the photometric term the model prefers
	to explain image evidence using lighting, rather than albedo,
	which leads to wrong skin tone predictions.
	Finally, replacing the weak-perspective camera with a perspective model
	would make the model more robust to extreme viewing angles and perspective
	distortion effects.
	Future work should look into denser forms of supervision,
	formulating a better photometric term and integrating a 
	perspective camera
	to resolve these issues.
	}
	}
    \label{fig:failures}
\end{figure*}